\documentclass[sn-mathphys,iicol]{sn-jnl}

\usepackage{lineno}
\usepackage{url} 
\usepackage{graphicx}
\usepackage[utf8]{inputenc}
\usepackage{amsmath}
\usepackage{amsfonts}
\usepackage{amssymb}
\usepackage{tabularx}
\usepackage{booktabs}
\usepackage{mathtools}
\usepackage{algpseudocode}
\usepackage{caption}

\DeclareMathOperator{\E}{\mathbb{E}}
\DeclareMathOperator{\U}{\mathbf{U}}
\DeclareMathOperator{\DKL}{\text{D}_\text{KL}}

\jyear{2022}%

\theoremstyle{thmstyleone}%
\theoremstyle{thmstyletwo}%

\theoremstyle{thmstylethree}%

\begin{document}

\title[Hierarchically Structured Task-Agnostic Continual Learning]{Hierarchically Structured Task-Agnostic Continual Learning}


\author*[1]{\fnm{Heinke} \sur{Hihn}}\email{heinke.hihn@uni-ulm.de}

\author[1]{\fnm{Daniel A.} \sur{Braun}}\email{daniel.braun@uni-ulm.de}

\affil[1]{\orgdiv{Institute of Neural Information Processing}, \orgname{Ulm University}, \orgaddress{\city{Ulm}, \country{Germany}}}


\abstract{
One notable weakness of current machine learning algorithms is the poor ability of models to solve new problems without forgetting previously acquired knowledge. The Continual Learning paradigm has emerged as a protocol to systematically investigate settings where the model sequentially observes samples generated by a series of tasks. In this work, we take a task-agnostic view of continual learning and develop a hierarchical information-theoretic optimality principle that facilitates a trade-off between learning and forgetting. We derive this principle from a Bayesian perspective and show its connections to previous approaches to continual learning. Based on this principle, we propose a neural network layer, called the Mixture-of-Variational-Experts layer, that alleviates forgetting by creating a set of information processing paths through the network which is governed by a gating policy. Equipped with a diverse and specialized set of parameters, each path can be regarded as a distinct sub-network that learns to solve tasks. To improve expert allocation, we introduce diversity objectives, which we evaluate in additional ablation studies. Importantly, our approach can operate in a task-agnostic way, i.e.,\ it does not require task-specific knowledge, as is the case with many existing continual learning algorithms. Due to the general formulation based on generic utility functions, we can apply this optimality principle to a large variety of learning problems, including supervised learning, reinforcement learning, and generative modeling. We demonstrate the competitive performance of our method on continual reinforcement learning and variants of the MNIST, CIFAR-10, and CIFAR-100 datasets.
}



\maketitle

\section{Introduction}
Acquiring new skills and concepts without forgetting previously acquired knowledge is a hallmark of human and animal intelligence. Biological learning systems leverage task-relevant knowledge from preceding learning episodes to guide subsequent learning of new tasks to accomplish this. Artificial learning systems, such as neural networks, usually lack this crucial property and experience a problem coined "catastrophic forgetting" \cite{mccloskey1989catastrophic}. Catastrophic forgetting occurs when we naively apply machine learning algorithms to solve a sequence of tasks $T_{1:t}$, where the adaptation to task $T_t$ prompts overwriting of the parameters learned for tasks $T_{1:t-1}$.

The Continual Learning (CL) paradigm \cite{thrun1998lifelong} has emerged as a way to investigate such problems systematically. We can divide CL approaches into four broad categories: generative approaches with memory consolidation, regularization, architecture and expansion methods, and algorithm-based methods. Generative methods train a generative model to learn the data-generating distribution to reproduce data of old tasks. Data sampled from the learned model is then part of the training process \cite{shin2017continual,rebuffi2017icarl}. This strategy draws inspiration from neuroscience research regarding the reactivation of neuronal activity patterns representing previous experiences that are hypothesized to be vital for stabilizing new memories while retaining old memories \cite{wilson1994reactivation,rasch2007maintaining,van2016hippocampal}. In contrast, regularization methods \cite{kirkpatrick2017overcoming,zenke2017continual,ahn2019uncertainty,benavides2020towards,han2021continual} introduce an additional constraint to the learning objective. The goal is to prevent changes in task-relevant parameters, where we may measure relevance as, e.g., the performance on previously seen tasks \cite{kirkpatrick2017overcoming,zenke2017continual,li2021lifelong,cha2020cpr}. This approach can also be motivated through synaptic plasticity and elasticity changes of biological neurons when learning new tasks \cite{ostapenko2019learning}. CL can also be achieved by modifying the design of a model during learning \cite{lin2019conditional,fernando2017pathnet,rusu2016progressive,yoon2018lifelong,golkar2019continual}. Such methods include adding new layers to a neural network \cite{zacarias2018sena}, re-routing data through a neural network based on task information \cite{collier2020routing}, distilling parameters \cite{zhai2019lifelong,liu2020mnemonics}, and adding new experts to a mixture-of-experts architecture \cite{Lee2020A}. Lastly, algorithmic approaches aim to adapt the optimization algorithm itself to avoid catastrophic forgetting, e.g., by mapping the gradient updates into a different space \cite{zeng2019continual,wang2021training}. Some methods provide a combination of approaches, as it seems plausible that a mixture of approaches will provide the best-performing systems, as these methods are often complementary \cite{biesialska2020continual,de2021continual,vijayan2021continual}.

\begin{figure*}[t!]
\centering
\begin{minipage}{0.49\textwidth}
\includegraphics[width=\textwidth, trim={0cm 1cm 0cm 0cm}, clip]{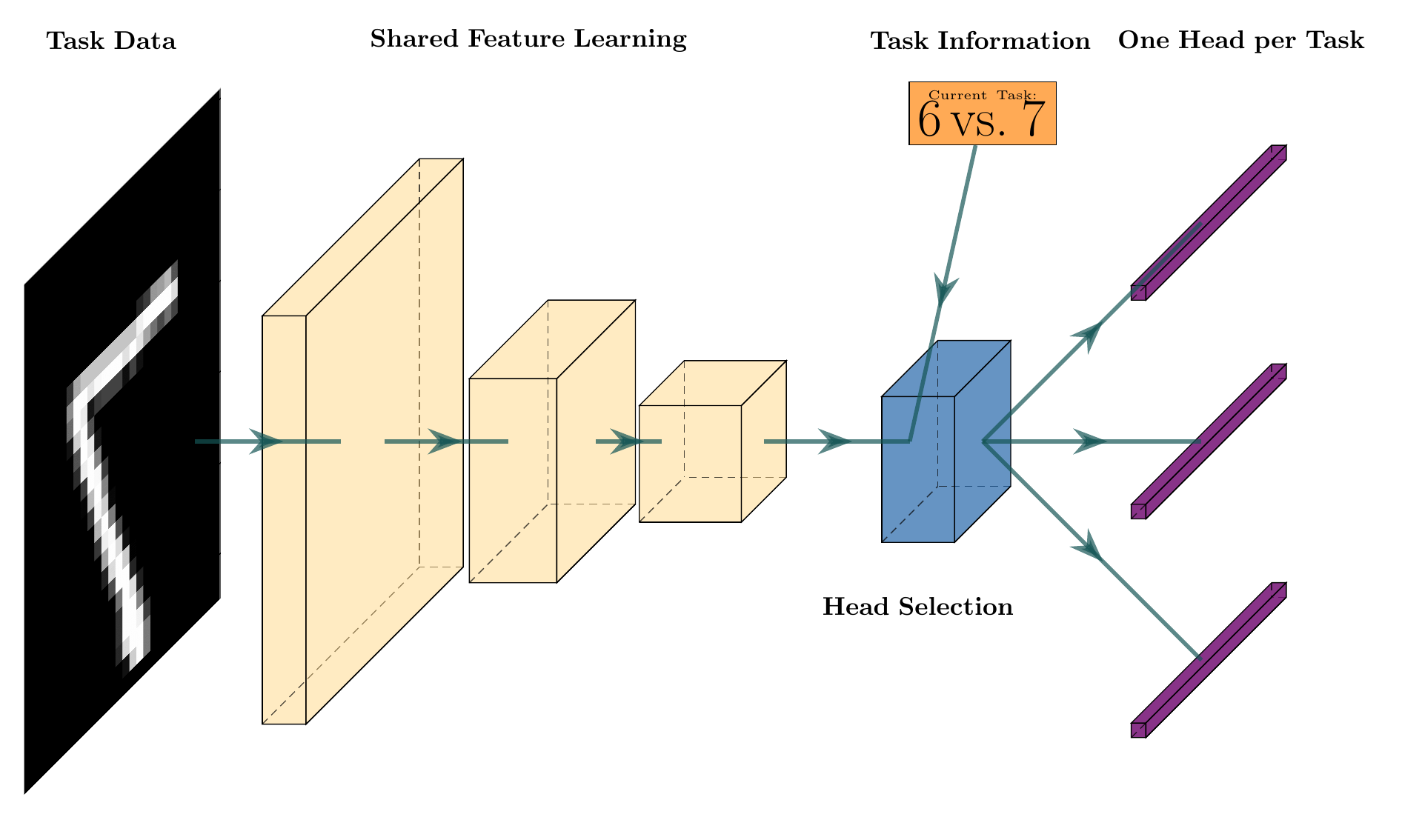}
\end{minipage}
\begin{minipage}{0.49\textwidth}
\includegraphics[width=\textwidth, trim={0cm 0cm 0cm 0cm}, clip]{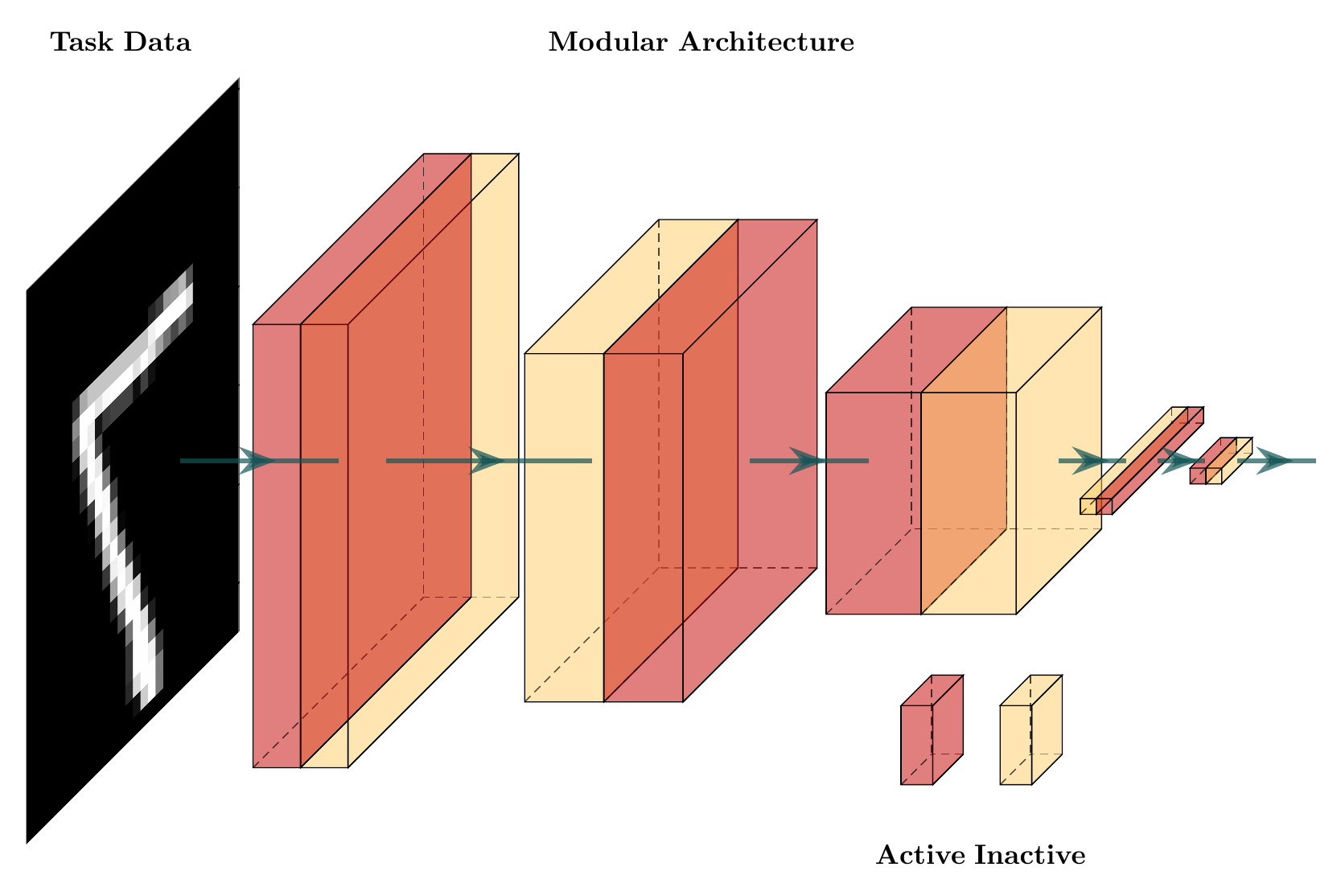}
\end{minipage}
\caption{Left: Multi-head architectures share a feature extractor block and select task-specific output heads based on the knowledge provided by a task oracle. Right: Our proposed architecture for single head CL. Each layer implements a modular system, such that task data can flow through distinct ''paths'' (highlighted) in the network.}
\label{fig:cl}
\end{figure*}

While there has been significant progress in the field of CL, there are still some major open questions \cite{parisi2019continual}. For example, most existing algorithms share a significant drawback in that they require task-specific knowledge, such as the number of tasks and which task is currently at hand \cite{zenke2017continual,shin2017continual,nguyen2017variational,kirkpatrick2017overcoming,li2017learning,rao2019continual,
 sokar2021self,yoon2018lifelong,han2021contrastive,chaudhry2021using,chaudhry2018efficient}. One prominent class of CL approaches sharing this drawback are multi-head approaches \cite{khatib2019strategies,nguyen2017variational,ahn2019uncertainty}, which build a set of shared layers but a separate output layer ("head") per task, deterministically activated by the current task index (see Figure \ref{fig:cl}).

Extracting relevant task information is in general a difficult problem, in particular when distinguishing tasks without any contextual input  \cite{hihn2020specialization,yao2019hierarchically}. Thus, providing the model with such task-relevant information yields overly optimistic results \cite{chaudhry2018riemannian}. In order to deal with more realistic and challenging CL scenarios, therefore, models must learn to compensate for the lack of auxiliary information. The approach we propose in this work tackles this problem by formulating a hierarchical learning system, that allows us to learn a set of sub-modules specialized in solving particular tasks. To this end, we introduce hierarchical variational continual learning (HVCL) and devise the mixture-of-variational-experts layer (MoVE layers) as an instantiation of HVCL. MoVE layers consist of $M$ experts governed by a gating policy, where each expert maintains a posterior distribution over its parameters alongside a corresponding prior. During each forward pass, the gating policy selects one expert per layer. This sparse selection reduces computation as only a small subset of the parameters must be updated during the back-propagation of the loss \cite{Shazeer2017}. To mitigate catastrophic forgetting we condition the prior distributions on previously observed tasks and add a penalty term on the Kullback-Leibler-Divergence between the expert posterior and its prior. This constraint facilitates a trade-off between learning and forgetting and allows us to design information-efficient decision-makers \cite{hihn2020specialization}.

When dealing with ensemble methods, two main questions arise naturally. The first one concerns the question of optimally selecting ensemble members using appropriate selection and fusion strategies \cite{Kuncheva2004}. The second one, is the question of how to ensure expert diversity \cite{kuncheva2003measures,bian2021when}. We argue that ensemble diversity benefits continual learning and investigate two complementary diversity objectives: the entropy of the expert selection process and a similarity measure between different experts based on Wasserstein exponential kernels in the context of determinantal point processes \cite{kulesza2012determinantal}. By maximizing the determinant of the expert similarity matrix, we can then ``spread'' the expert parameters optimally within the shared parameter space. To summarize, our contributions are the following: $(i)$ we extend variational continual learning \cite{nguyen2017variational} to a hierarchical multi-prior setting, $(ii)$ we derive a computationally efficient method for task-agnostic continual learning from this general formulation, $(iii)$ to improve expert specialization and diversity, we introduce and evaluate novel diversity measures $(iv)$ we demonstrate our approach in supervised CL, generative CL, and continual reinforcement learning. 

This paper is structured as follows: after introducing our method in Section \ref{sec:clmove}, we design, perform, and evaluate the main experiments in Section \ref{sec:experiments}. In Section~\ref{sec:discussion}, we discuss novel aspects of the current study in the context of previous literature and conclude with a final summary in Section \ref{sec:conclusion}.

\section{Hierarchical Variational Continual Learning}
\label{sec:clmove}
\label{sec:hvcl}
In this section we first introduce the concept of continual learning and related nomenclature formally and then extend the variational continual learning (VCL) setting introduced by Nguyen et al.\ \cite{nguyen2017variational} to a hierarchical multi-prior setting and then introduce a neural network implementation as a generalized application of this paradigm in Section~\ref{sec:sparse}.

\begin{figure*}
\centering
\includegraphics[width=0.9\textwidth, trim={1cm 21cm 4cm 1cm}, clip]{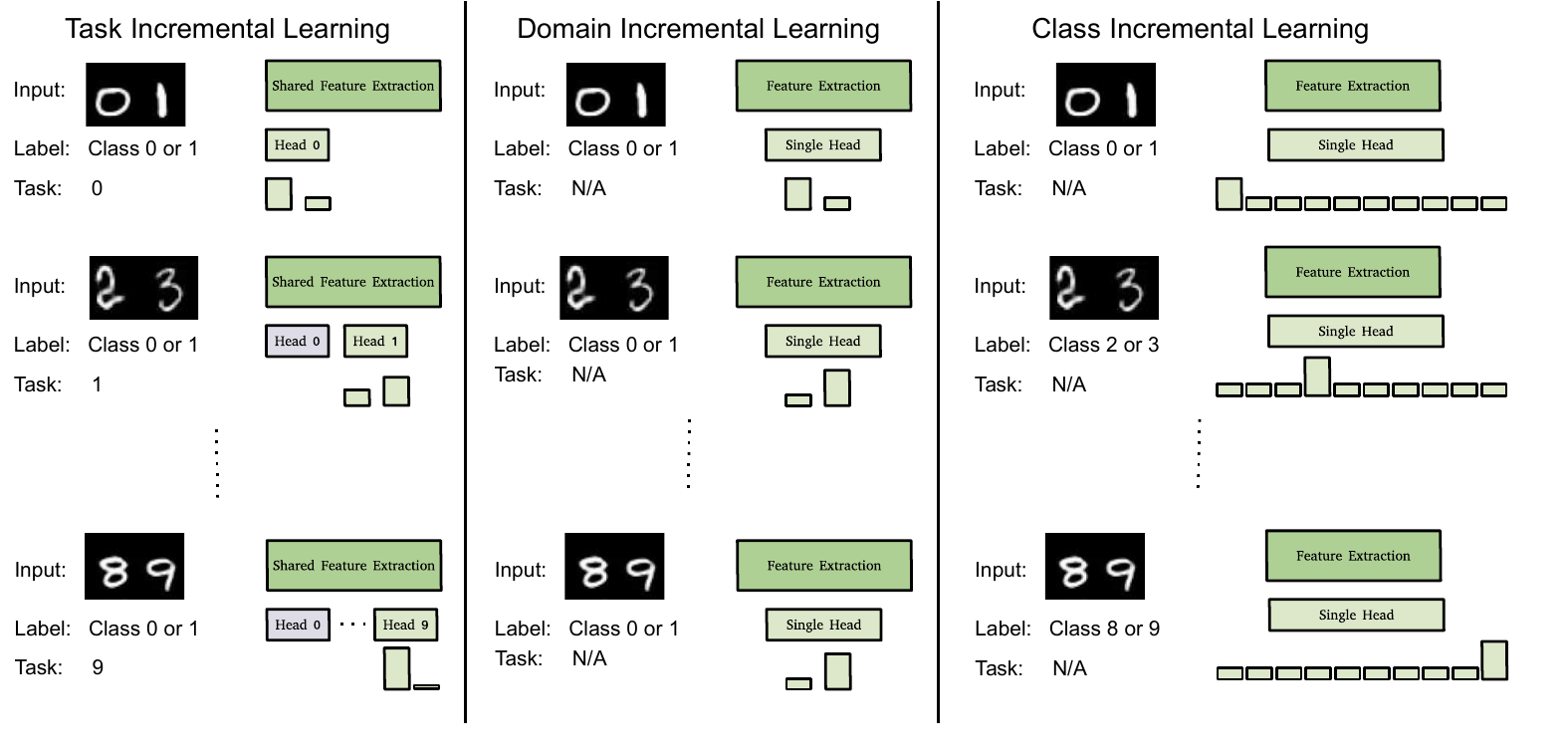} 
\caption{Continual Learning setups. In Task Incremental Learning the number of classes is constant and task information is provided. In Domain Incremental Learning, the number of classes is also constant, but no task information is given. In Class Incremental Learning the total number of classes is constant, but the learning agent only observes a subset at a time, e.g., two out of ten total classes per task.}
\label{fig:clsetup}
\end{figure*}

In CL the goal is to minimize the loss of all seen tasks given no (or limited) access to data from previous tasks:
\begin{equation}
\min \sum_{t=1}^T \mathbb{E}_{(x_t,y_t)\thicksim (X_t,Y_t)}\left[ \ell(f_\theta(x_t, y_t)) \right],
\end{equation}
where $T$ is the total number of tasks, $\mathcal{D}_t = \{x_t^i, y_t^i\}_{i=1}^{N_t} = (X_t, Y_t)$ the dataset of task $t$, $\ell$ some loss function, and $f_\theta$ a \textit{single} predictor parameterized by $\theta$ (e.g.,\ a neural network). Here, task refers to an isolated training stage with a new dataset. This dataset may belong to a new set of classes, a new domain, or a new output space. We can divide these concepts based on the in- and output distributions \cite{de2021continual}: in task-incremental learning the labels change ${Y_t} \neq {Y_{t+1}}$, but the label distributions remain $P(Y_t) = P(Y_{t+1})$, e.g.,\ in a series of binary classification tasks, plus the task $t$ is always known. In domain-incremental learning we have ${Y_t} = {Y_{t+1}}$ and $P(Y_t) = P(Y_{t+1})$, but $t$ is unknown. Finally, in class-incremental learning we have ${Y_t} \subset {Y_{t+1}}$, but $P(Y_t) \neq P(Y_{t+1})$ and unknown $t$, e.g.,\ MNIST with ten total classes but only two classes per task. For all settings it holds that $X_t \neq X_{t+1}$ -- see Figure \ref{fig:clsetup} for an illustration.

The variational continual learning approach  \cite{nguyen2017variational} describes a general learning paradigm wherein an agent stays close to an old strategy ("prior") it has learned on a previous task $T_{t-1}$ while learning to solve a new task $T_t$ ("posterior"). 
Given datasets of input-output pairs $\mathcal{D}_t = \{x_t^i, y_t^i\} _{i=0}^{N_t}$  of tasks $t \in \{1, ..., T\}$, the main learning objective of minimizing the log-likelihood $\log p_\theta(y_t^{i}\vert x_t^{i})$ for task $t$ is augmented with an additional loss term in the following way:
\begin{equation}
\label{eq:vcl}
\begin{aligned}
\mathcal{L}_{\text{VCL}}^t = \sum_{i=1}^{N_t} \mathbb{E}_{\theta}\left[\log p_\theta(y_t^{i}\vert x_t^{i})\right] - \DKL\left[p_t(\theta)\vert \vert p_{t-1}(\theta) \right],
\end{aligned}
\end{equation}
where $p(\theta)$ is a distribution over the models parameter $\theta$ and $N_t$ is the number of samples for task $t$.
The prior constraint encourages the agent to find an optimal trade-off between solving a new task and retaining knowledge about old tasks. When the likelihood model is implemented by a neural network, a new output layer can be associated with each incoming task, resulting in a multi-head implementation. Over the course of $T$ datasets, Bayes' rule then recovers the posterior
\begin{align}
\label{eq:bayesinference}
p(\theta\vert \mathcal{D}_{1:T}) & \propto p(\theta) \prod_{t=1}^{T}\prod_{i=1}^{N_t}p(y_t^{i}\vert \theta,x_t^{i}) \nonumber \\ 
& = p(\theta)\prod_{t=1}^T p(\mathcal{D}_t\vert \theta) \\ 
& \propto p(\theta\vert \mathcal{D}_{1:T-1})p(\mathcal{D}_T\vert \theta),
 \nonumber
\end{align}
which forms a recursion: the posterior after seeing $T$ datasets is obtained by multiplying the posterior after $T-1$ with the likelihood and normalizing accordingly.

In their original work, the authors propose to use multi-headed networks, i.e.,\ to train a new output layer for each incoming task. This strategy has two main drawbacks: $(i)$ it introduces an organizational overhead due to the growing number of network heads, and $(ii)$ task boundaries must be known at all times, making it unsuitable for more complex continual learning settings. In the following we argue that we can alleviate these problems by combining multiple decision-makers with a learned selection policy to replace the deterministic head selection. 

To extend VCL to the hierarchical case, we  assume that samples are drawn from a set of $M$ independent data generating processes, i.e.\ the likelihood is given by a mixture model $p(y\vert x) = \sum_{m=1}^M p(m\vert x)p(y\vert m,x)$. We define an indicator variable $z \in Z$, where $z_m^{i,t}$ is $1$ if the output $y_i^t$ of sample $i$ from task $t$ was generated by expert $m$ and zero otherwise. The conditional probability of an output is then given by
\begin{equation}
\label{eq:mixturelikelihood}
p(y_t^i\vert x_t^i,\Theta) = \sum_{m=1}^M p(z_t^{i,m}\vert x_t^i,\vartheta)p(y_t^i\vert x_t^i,\omega_m),
\end{equation}
where $\vartheta$ are the parameters of the selection policy, $\omega_m$ the parameters of the $m$-th expert, and $\Theta = \{\vartheta, \{\omega_m\}_{m=1}^M\}$ the combined model parameters. The posterior after observing $T$ tasks is then given by
\begin{align}
\label{eq:hierarchicalbayesinference}
p(\Theta\vert \mathcal{D}_{1:T}) & \propto p(\vartheta)p(\omega) \prod_{t=1}^{T}\prod_{i=1}^{N_t} \sum_{m=1}^M p(z_t^{i,m}\vert x_t^i,\vartheta)p(y_t^i\vert x_t^i,\omega_m)  \nonumber \\
& =  p(\Theta)\prod_{t=1}^T p(\mathcal{D}_t\vert \Theta) \\ 
& \propto p(\Theta\vert \mathcal{D}_{1:T-1})p(\mathcal{D}_T\vert \Theta). \nonumber
\end{align}
The Bayes posterior of an expert $p(\omega_m\vert \mathcal{D}_{1:T})$ is recovered by computing the marginal over the selection variables $Z$. Again, this forms a recursion, in which the posterior $p(\Theta\vert \mathcal{D}_{1:T})$ depends on the posterior after seeing $T-1$ tasks and the likelihood $p(\mathcal{D}_T\vert \Theta)$. We can now formulate the hierarchical variational continual learning objective for task $t$ as minimizing the following loss:
\begin{equation}
\label{eq:hvcl}
\begin{aligned}
\mathcal{L}_{\text{HVCL}}^t & = \sum_{i=1}^{N_t} \mathbb{E}_{p(\Theta)}\left[\log p(y_t^{i}\vert x_t^{i}, \Theta)\right] \\ 
& - \DKL\left[p_t(\vartheta)\vert \vert p_{1:t-1}(\vartheta)\right] \\ 
& - \DKL\left[p_t(\omega)\vert \vert p_{1:t-1}(\omega)\right],
\end{aligned}
\end{equation}
where $N_t$ is the number of samples in task $t$, and the likelihood $p(y\vert x, \Theta)$ is defined as in Equation~\eqref{eq:mixturelikelihood}. The Mixture-of-Variational-Experts layers we introduce in Section~\ref{sec:sparse} are based on a generalization of this optimization problem.

\subsection{Sparsely Gated Mixture-of-Variational Layers}
\label{sec:sparse}

\begin{figure*}[t!]
\centering
\includegraphics[width=0.95\textwidth, trim={16cm 0.5cm 3.5cm .5cm}, clip]{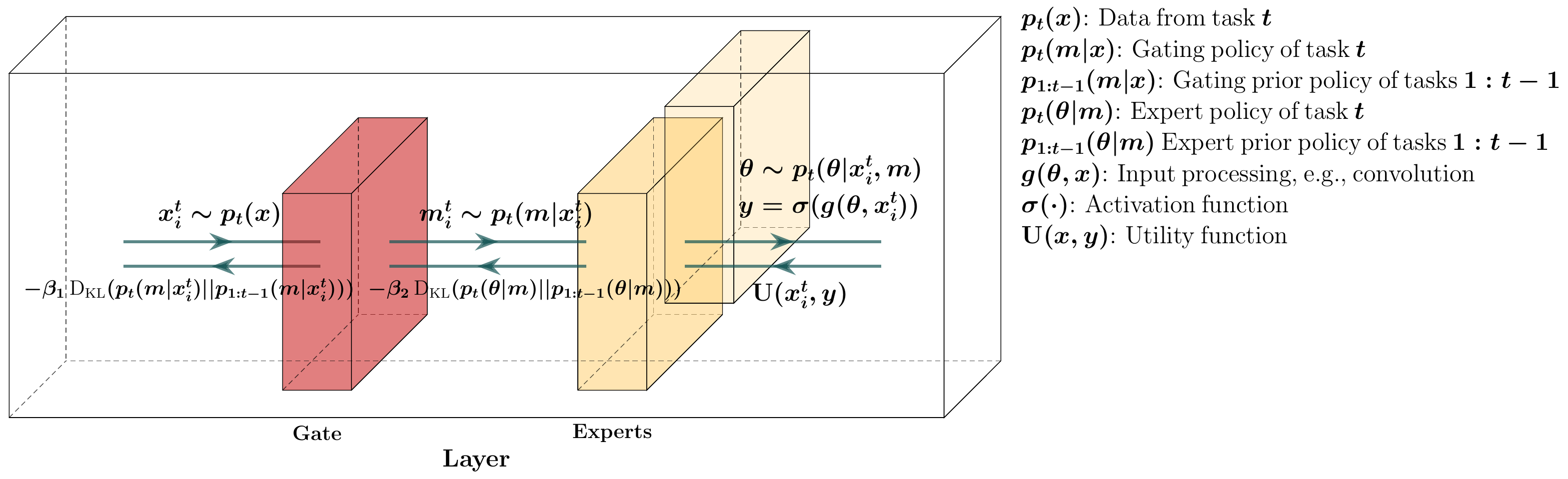}
\includegraphics[width=0.95\textwidth, trim={3cm 1.25cm 1.25cm 0.5cm}, clip]{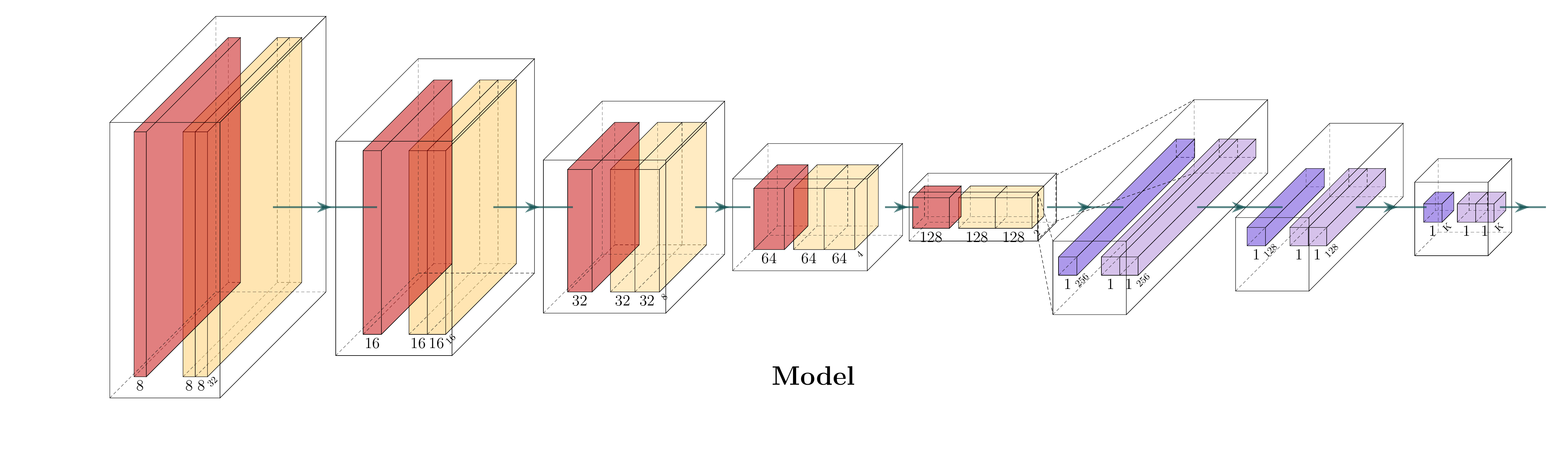}
\caption{This figure illustrates our proposed design. Each layer implements a top-$k$ expert selection conditioned on the output of the previous layer. Each expert $m$ maintains a distribution over its weights $p(\theta\vert m) = \mathcal{N}(\mu_m, \sigma_m)$ and a set of bias variables $b_{m}$. Left to right arrows represent sampling, while right to left arrows show the resulting utility and losses. Combining $L$ layers with $M$ experts gives $L^M$ possible paths through the network. The architecture shown here is used in the CIFAR10 experiments.}
\label{fig:architecture}
\end{figure*}

As we plan to tackle not only supervised learning problems, but also reinforcement learning problems, we assume in the following a generic scalar utility function $\U(x,f_\theta(x))$ that depends both on the input $x$ and the parameterized agent function
$f_\theta(x)$ that generates the agent's output $y$. We assume that the agent function $f_\theta(x)$ is composed of multiple layers as depicted in 
Figure~\ref{fig:architecture}.
Our layer design builds on the sparsely gated Mixture-of-Expert (MoE) layers \cite{Shazeer2017}, which in turn draws on the Mixture-of-Experts paradigm introduced by Jacobs et al.\ \cite{Jacobs1991}. MoEs consist of a set of $M$ experts indexed by $m$ and a gating network $p(m\vert x)$ whose output is a (sparse) $M$-dimensional vector. All experts have an identical architecture but separate parameters. Let $p(m\vert x)$ be the gating output and $p(y\vert m,x)$ the response of an expert $m$ given input $x$. The layer's output is then given by a weighted sum of the experts responses, i.e.,~ $p(y\vert x) = \sum_{m \in M} p(m\vert x)p(y\vert m,x)$. To save computation time we employ a top-$k$ gating scheme, where only the $k$ experts with highest gating activation are evaluated and use an additional penalty that encourages gating sparsity (see Section \ref{sec:sparse}). In all our experiments we set $k = 1$, to drive expert specialization (see Section \ref{sec:specdiv}) and reduce computation time.
We implement the learning objective for task $t$ as layer-wise regularization in the following way:
\begin{equation}
\begin{aligned}
\mathcal{L}_{\text{MoVE}}^{t} & = \sum_{i=1}^{N_t} \bigg[\mathbb{E}_{\Theta}\left[\U(x_i^t, f_\Theta(x_i^t))\right] \\ 
& - \sum_{l=1}^{L} \mathbb{E}_{p^l(m\vert x_i^t), q^l(\theta\vert m)} \bigg[ \\
& \beta_1 \DKL\left[p_t^l(m\vert x_i^t)\vert \vert p_{1:t-1}^l(m\vert x_i^t)\right] \\
&  + \beta_2 \DKL\left[p_t^l(\theta\vert m)\vert \vert p_{1:t-1}^l(\theta\vert m) \right] \bigg] \bigg],
\end{aligned}
\label{eq:move}
\end{equation}
where $L$ is the total number of layers,  $\Theta = \{\theta, \{\vartheta_m\}_{m=1}^M\}$ the combined parameters,and the temperature parameters $\beta_1$ and $\beta_2$ govern the layer-wise trade-off between utility (e.g.,\ classification performance) and information-cost. 

Thus, we allow for two major generalizations compared to Equation \ref{eq:hvcl}: in lieu of the log-likelihood we allow for generic utility functions $\U(\cdot)$, and instead of applying the constraint on the gating parameters, we apply it directly on the gating output distribution $p(m\vert x)$. This implies, that the weights of the gating policy are not sampled. Otherwise the gating mechanism would involve two stochastic steps: one in sampling the weights and a second one in sampling the experts. This potentially high selection variance hinders expert specialization (see Section \ref{sec:specdiv}). Encouraging the gating policy to stay close to its prior also ensures that similar inputs are assigned to the same expert. Next we consider how we could extend objective \ref{eq:move} further by additional terms that encourage diversity between different experts.

\subsection{Encouraging Expert Diversity}
\label{subsec:diversity}
In the following, we argue that a diverse set of experts may mitigate catastrophic forgetting in continual learning, as experts specialize more easily in different tasks, which improves expert selection. Diversity measures enjoy an increasing interest in the reinforcement learning community but remain mainly understudied in continual learning  \cite[e.g.,\ ][]{bang2021rainbow}. In the reinforcement learning literature diversity has been considered, for example, by encouraging skills or policies that are  sufficiently different \cite{eysenbach2018diversity,parker2020effective}, or by  sampling trajectories that reflect goal diversity \cite{dai2021diversity}, which is an idea similar to well-known bagging techniques \cite{Breiman1996}. Moreover, diversity may arise from a sufficiently high variance weight initialization, but this can introduce computational instabilities during back-propagation, as we lose the variance reducing benefits of state-of-the-art initialization schemes \cite{glorot2010understanding,he2015delving,narkhede2021review}. Also, there is no guarantee that the expert parameters won't collapse again during training. 

In the following, we present two expert diversity objectives. The first one arises directly from the main learning objective and is designed to act as a regularizer on the gating policy while the second one is a more sophisticated approach that aims for diversity in the expert parameter space. The latter formulation introduces a new class of diversity measures, as we discuss in more detail in Section \ref{sec:related}. We designed additional experiments in Section \ref{sec:ablation} to investigate their influence on learning and the resulting policies and to further motivate the need for expert diversity.

\subsubsection{Diversity through Specialization}
\label{sec:specdiv}
The relationship between objectives of the form described by Equation~\eqref{eq:move} 
with the emergence of expert specialization has been previously investigated for simple learning problems \cite{Genewein2015} and in the context of meta-learning \cite{hihn2020specialization}, but not in the context of continual learning.
This class of models assumes a two-level hierarchical system of specialized decision-makers where first level decision-makers $p(m\vert x)$ select which second level decision-maker $p(y\vert m,x)$ serves as experts for a particular input $x$. By co-optimizing 
\begin{equation}
\label{eq:par_mutual}
\max_{p(y\vert x,m), p(m\vert x)} \E[\mathbf{U}(x,y)] - \beta_1 I(X;M) - \beta_2 I(X;Y\vert M),
\end{equation}
the combined system finds an optimal partitioning of the input space $X$, where $I(\cdot\vert \cdot)$ denotes the (conditional) mutual information between random variables. 
In fact, the hierarchical VCL objective given by Equation \eqref{eq:hvcl} can be regarded as a special case of the information-theoretic objective given by Equation \eqref{eq:par_mutual}, if we interpret the prior as the learning strategy of task $t-1$ and the posterior as the strategy of task $t$, and set $\beta = 1$. All these hierarchical decision systems correspond to a multi prior setting, where different priors associated with different experts can specialize on different sub-regions of the \emph{input space}. In contrast, specialization in the context of continual learning can be regarded as the ability of partitioning the \emph{task space}, where each expert decision-maker $m$ solves a subset of old tasks $T^m \subseteq T_{1:t}$. In both cases, expert diversity is a natural consequence of specialization if the gating policy $p(m\vert x)$ partitions between the experts. Using gradient descent on parameterized distributions, objective \eqref{eq:par_mutual} can also be maximized in an online manner \cite{hihn2020specialization}.

In addition to the implicit pressures for specialization already implied by Equation~\ref{eq:move}, here we investigate the effect of an additional entropy cost. Inspired by recent entropy regularization techniques \cite{eysenbach2018diversity,galashov2019information,grau2018soft}, we aim to improve the gating policy by introducing the entropy cost
\begin{equation}
\max_{p(m\vert x)} H(M\vert X) - H(M),
\label{eq:entropydiv}
\end{equation}
where $M$ is the set of experts and $X$ the inputs. By maximizing the conditional entropy $H(M\vert X) = -\sum_{m,x} p(x) p(m\vert x) \log p(m\vert x)$ we encourage high certainty in the expert gating and by minimizing the marginal entropy $H(M)$ we prefer solutions that minimize the number of active experts. In our implementation, we compute these values batch-wise, as the full entropies are not tractable. We evaluate these entropy penalties in Section \ref{sec:ablation}.

\subsubsection{Parameter Diversity}
\label{sec:paramdiv}
\begin{figure*}[t!]
\centering
\includegraphics[width=0.55\textwidth, trim={.25cm 27.75cm 15.5cm 0cm}, clip]{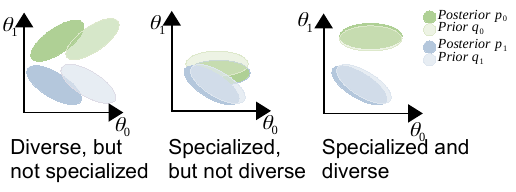}
\includegraphics[width=0.55\textwidth, trim={0cm 27cm 12.4cm 0cm}, clip]{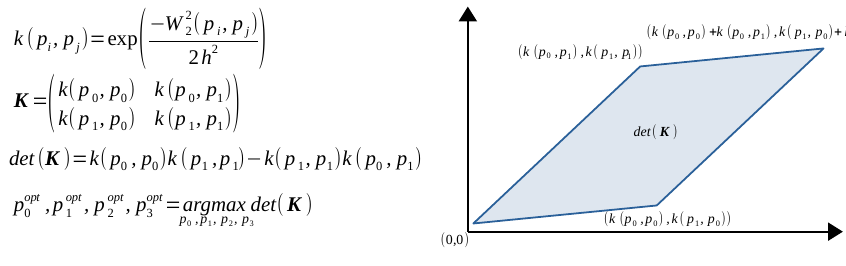}
\caption{Upper Row: We seek experts that are both specialized, i.e., their posterior $p$ is close to their prior $q$, and diverse, i.e., posteriors $p_i,p_j \forall i \neq j$ are sufficiently distant from one another. Bottom Row: To this effect, we maximize the determinant of the kernel matrix $K$, effectively filling the feature space. In the case of two experts this would mean to maximize $det(\textbf{K}) = 1 -  K(p_0,p_1)$, which we can achieve by maximizing the Wasserstein-2 distance between the posteriors $p_0$ and $p_1$. }
\label{fig:specialization_diversity}
\end{figure*}
Our second diversity formulation is based on differences in the parameter space, rather than in the output space as formalized by Equations \eqref{eq:entropydiv} or \eqref{eq:par_mutual}. To find expert parameters that are pairwise different, we introduce a symmetric distance measure and we show how this measure can be efficiently computed and maximized. By maximizing distance in parameter space, we hope to find a set of expert parameters stretched over the space of possible parameters. This in turn helps to prevent collapsing to a state where all experts have similar parameters (and thus similar outputs), rendering the idea behind an ensemble useless.

Our idea builds on determinantal point processes (DPPs) \cite{kulesza2012determinantal}, a mechanism that produces diverse subsets by sampling proportionally to the determinant of the kernel matrix of points within the subset \cite{macchi1975coincidence}. A point process $P$ on a ground set $Y$ is a probability measure over finite subsets of $Y$. A sample from $P$ may be the empty set, the entirety of $Y$, or anything in between. $P$ is a determinantal point process if, when $Y$ is a random subset drawn according to $P$, we have, for every$A \subset Y$, $P(A \subset Y) = \mathrm{det}(K_A)$ for some real, symmetric $N\times N$ matrix $K$ indexed by the elements of $Y$. Here, $K_A = [K_{ij}]_{i,j\in A}$ denotes the restriction of $K$ to the entries indexed by elements of $A$, and we adopt $\mathrm{det}(K_\emptyset) = 1$. Since $P$ is a probability measure, all principal minors $\mathrm{det}(KA)$ of $K$ must be non-negative, and thus $K$ itself must be positive semi-definite. These requirements turn out to be sufficient:  any $K$, $0\leq K \leq I$, defines a DPP. We refer to $K$ as the marginal kernel since it contains all the information needed to compute the probability of any subset $A$ of $Y$. If $A = \{i\}$ is a singleton, then we have $P(i\in Y) = K_{i,i}$. In this case, the diagonal of $K$ gives the marginal probabilities of inclusion for individual elements of $Y$.  Diagonal entries close to $1$ correspond to elements of $Y$ selected with high probability. The matrix $K$ is defined by a kernel function $k(x_0,x_1)$. A kernel is a two-argument real-valued function over $\mathcal{X}\times\mathcal{X}$ such that for any $x_0,x_1 \in \mathcal{X}$:
\begin{equation}
\label{eq:kernel}
k(x_0,x_1) = \langle\phi(x_0),\phi(x_1)\rangle_\mathcal{F},
\end{equation}
where $\mathcal{X}$ is a vector space and $\mathcal{F}$ is a inner-product space such that $\forall x \in \mathcal{X}: \phi(x) \in \mathcal{F}$. 
Specifically, we use a exponential kernel based on the Wasserstein-2 distance $W(p,q)$ between two probability distributions $p$ and $q$. The $p^\text{th}$ Wasserstein distance between two probability measures $p$ and $q$ in $P_p(M)$ is defined as
\begin{equation}
\label{eq:wasserstein}
W_p (p, q):=\left( \inf_{\gamma \in \Gamma (p, q)} \int_{M \times M} d(x, y)^p \, \mathrm{d} \gamma (x, y) \right)^{1/p},
\end{equation}
where $\Gamma(p,q)$ denotes the collection of all measures on $M \times M$ with marginals $p$ and $q$ on the first and second factors. Let $p$ and $q$ be two isotropic Gaussian distributions and $W_2^2(q,p)$ the Wasserstein-2 distance between $p$ and $q$. The exponential Wasserstein-2 kernel is then defined by 
\begin{equation}
\label{eq:w2expkernel}
k(p,q) = \exp\left(-\frac{W_2^2(p,q)}{2h^2}\right),
\end{equation}
where $h$ is the kernel width. We show in Appendix \ref{app:wasserkernel} that Equation \eqref{eq:w2expkernel} gives a valid kernel. This formulation has two properties that make it suitable for our purpose. Firstly, the Wasserstein distance is symmetric, i.e., $W_2^2(p,q) = W_2^2(q,p)$, which in turn will lead to a symmetric kernel matrix. This is not true for other similarity measures on probability distributions, such as $\DKL$ \cite{cover2012elements}. Secondly, if $p$ and $q$ are Gaussian and mean-field approximations, i.e., covariance matrices $\Sigma_p$ and $\Sigma_q$ are given by diagonal matrices, such that $\Sigma_p = \text{diag}(d_p)$ and $\Sigma_q = \text{diag}(d_q)$, $W_2^2(p,q)$ can be computed in closed form as 
\begin{equation}
\label{eq:gaussianwasser}
W_2^2(p,q) = \vert \vert \mu_p - \mu_q\vert \vert _2^2 + \vert \vert \sqrt{d_p} - \sqrt{d_q}\vert \vert _2,
\end{equation}
where $\mu_{p,q}$ are the means and $d{p,q}$ the diagonal entries of distributions $p$ and $q$. We provide a more detailed derivation of Equation \eqref{eq:gaussianwasser} in Appendix \ref{app:wassergaussian}. For each layer $l$ with $N$ experts the following regularization objective is added to the main objective:
\begin{equation}
\max \sum_{l=1}^L \mathrm{det} \begin{pmatrix}
1 & k(p_1^l, p^l_2) & \cdots & k(p^l_1, p_N^l) \\
k(p^l_2, p^l_1) & 1 & \cdots & k(p^l_2, p^l_N) \\
\vdots & \vdots & \ddots & \vdots \\
k(p^l_N, p^l_0) & k(p^l_N, p^l_2) & \cdots & 1 \\
\end{pmatrix}
\end{equation}
where $L$ is the number of Layers, $k(p^l_i,p^l_j)$ is the kernel of the $i$-th and $j$-th expert of layer $l$ (see Equation \ref{eq:kernel}) and $\mathrm{det}(K)$ denotes the matrix determinant of the kernel matrix $K$. Note that the matrix is symmetric, which reduces computation time. Computing $\mathrm{det}(K)$ can have some pitfalls, which we discuss in Section \ref{sec:future}.

From a geometric perspective, the determinant of the kernel matrix represents the volume of a parallelepiped spanned by feature maps corresponding to the kernel choice. We seek to maximize this volume, effectively filling the parameter space -- see Figure \ref{fig:specialization_diversity} for an illustration.

\section{Experiments}
\label{sec:experiments}
\begin{table*}[t!]
\small
\setlength{\tabcolsep}{2pt} 
\centering
\begin{tabularx}{\textwidth}{*{3}l}
\toprule
\textbf{Baselines} &  \textbf{S-MNIST} & \textbf{P-MNIST} \\
\midrule
Dense Neural Network &  86.15 ($\pm$1.00)  & 17.26 ($\pm$0.19) \\
Offline re-training &  99.64 ($\pm$0.03)& 97.59 ($\pm$0.02) \\
\midrule
\textbf{Single-Head and Task-Agnostic Methods} & & \\
\midrule
Hierarchical VCL (ours) & 97.50 ($\pm0.33$) & 97.07 ($\pm$0.62) \\
Hierarchical VCL w/ GR (ours) & 98.60 ($\pm0.35$) & 97.47 ($\pm$0.52) \\
Uncertainty Guided CL w/ BNN \cite{ebrahimi2020uncertainty} & 97.70 ($\pm$0.03) & 92.50 ($\pm$0.01) \\
Brain-inspired Replay through Feedback$^\dagger$ \cite{van2020brain} & 99.66 ($\pm$0.13) & 97.31 ($\pm$0.04) \\
Hierarchical Indian Buffet Neural Nets \cite{kessler2021hierarchical} & 91.00 ($\pm$2.20) & 93.70 ($\pm$0.60)\\
Balanced Continual Learning \cite{raghavan2021formalizing} & 98.71 ($\pm$0.06) & 97.51 ($\pm$0.05) \\
Target Layer Regularization \cite{mazur2021target} & 80.64 ($\pm$1.25) &\\
\bottomrule
\end{tabularx}
\caption{Continual learning results in the split MNIST (S-MNIST) and permuted MNIST (P-MNIST) benchmark compared to current CL methods. Results were averaged over ten random seeds with the standard deviation given in parenthesis. Results on algorithms marked with $\dagger$ were taken from \cite{van2018generative}, others from their original work. ''Dense Neural Network'' refers to simple NN, that has been trained naively with sequential data and represents a lower bound. ''Offline re-training'' refers to a NN that has been retrained on all tasks seen so far.}
\label{tab:mnistresults}
\end{table*}
While the correct and robust evaluation of continual learning algorithms is still a topic of discussion \cite{farquhar2018towards,hsu2018re}, we follow the majority of studies to ensure a fair comparison. We evaluate our approach in current supervised learning benchmarks in Section \ref{sec:scenarios}, in a generative learning setting in Section \ref{subsec:vae}, and in the continual reinforcement learning setup in Section \ref{sec:crl}. Additionally, we conduct ablation studies in Section \ref{sec:ablation} to investigate the influence of the diversity objective, the generator quality, the influence of the hyper-parameters $\beta_1$ and $\beta_2$, and regarding the number of experts. We give experimental details in Appendix \ref{app:cifarexperiments}.

\subsection{Continual Supervised Learning Scenarios}
\label{sec:scenarios}

The basic setting of continual learning is defined as an agent which sequentially observes data from a series of tasks $\mathcal{T} = \{T_i\}_{i=1}^N$ and must learn $T_i$ while maintaining performance on older tasks $T_{\leq i}$. We evaluate the performance of our method in this setting in split MNIST (Figure \ref{tab:mnistresults}), split CIFAR-10 and split CIFAR-100 (Figure \ref{tab:cifarresults}). We follow the domain incremental setup \cite{van2020brain,kessler2021hierarchical,raghavan2021formalizing,hsu2018re,mazur2021target,he2022online}, where the number of classes is constant and task information is not available, but we also compare against task-incremental methods
 \cite{zenke2017continual,shin2017continual,nguyen2017variational,kirkpatrick2017overcoming,li2017learning,rao2019continual,
 sokar2021self,yoon2018lifelong,han2021contrastive,chaudhry2021using,chaudhry2018efficient} in Appendix \ref{app:cifarexperiments} Tables \ref{tab:appmnistresults} and \ref{tab:appcifarresults}, where the task information is available, to give a complete overview of current methods. The different Continual Learning setups are given in Figure \ref{fig:clsetup} in more detail. 

The first benchmark builds on the MNIST dataset. Five binary classification tasks from the MNIST dataset arrive in sequence: 0/1, 2/3, 4/5, 6/7, and 8/9. In time step $t$, the performance is measured as the average classification accuracy on all tasks up to task $t$. In permuted MNIST the task received at each time step $t$ consists of labeled MNIST images whose pixels have undergone a fixed random permutation. The second benchmark is a variation of the CIFAR-10/100 datasets. In Split CIFAR-10, we divide the ten classes into five binary classification tasks. In total, CIFAR-10 consists of 60000 $32\times 32\times 3$ images in 10 classes, with 6000 images per class. There are 50000 training images and 10000 test images. CIFAR-100 is like the CIFAR-10, except it has 100 classes containing 600 images each. There are 500 training images and 100 testing images per class. Tasks are defined as a 10-way classification problem, thus forming ten tasks in total. 

We achieve comparable results to current state-of-the-art approaches (see Table~\ref{tab:mnistresults} and \ref{tab:cifarresults}) on all three supervised learning benchmarks. 
\subsection{Generative Continual Learning}
\label{subsec:vae}
\begin{table*}[t!]
\small
\setlength{\tabcolsep}{2pt} 
\centering
\begin{tabularx}{\textwidth}{*{3}l}
\toprule
\textbf{Baselines} &  \textbf{Split-CIFAR-10} & \textbf{CIFAR-100} \\
\midrule
Conv. Neural Network &  66.62 ($\pm$1.06)  & 19.80 ($\pm$0.19) \\
Offline re-training & 80.42 ($\pm$0.95)& 52.30 ($\pm$0.02) \\
\midrule
\textbf{Single-Head and Task-Agnostic Methods} & & \\
\midrule
Hierarchical VCL (ours) & 78.41 ($\pm$1.18) & 33.10 ($\pm$0.62) \\
Hierarchical VCL w/ GR (ours) & 81.00 ($\pm$1.15) & 37.20 ($\pm$0.52) \\
Continual Learning with Dual Regularizations \cite{han2021continual} & 86.72 ($\pm$0.30) & 25.62 ($\pm$0.22) \\
Natural Continual Learning \cite{kao2021natural} & & 38.79 ($\pm$0.24) \\ 
Target Layer Regularization \cite{mazur2021target} & 74.89 ($\pm$0.61 & \\
Memory Aware Synapses \cite{he2022online} & 73.50 ($\pm$1.54)  & \\
\bottomrule
\end{tabularx}
\caption{Continual learning results in the split CIFAR-10 and the split CIFAR-100 benchmark compared to current CL methods. Results were averaged over ten random seeds with the standard deviation given in the parenthesis. We report results of other methods as given in their original studies. ''Conv. Neural Network'' refers to simple CNN, that has been trained naively with sequential data and represents a lower bound. ''Offline re-training'' refers to a CNN that has been retrained on all tasks seen so far.}
\label{tab:cifarresults}
\end{table*}
Generative CL is a simple but powerful paradigm \cite{shin2017continual,rebuffi2017icarl,van2020brain}. The main idea is to learn the data generating distribution and simulate data of previous tasks. We can extend our approach to the generative setting by modeling a variational autoencoder using the novel layers we propose in this work. We provide hyper-parameters and other experimental settings in Appendix \ref{app:cifarexperiments}.

Modeling the latent variable $z$ to capture the dynamics the data generating distribution $p(x)$ is difficult if $p(x)$ is multi-modal and authors have suggested the use of more complex distributions \cite{hadjeres2017glsr,ghosh2019variational,vahdat2020nvae} as variational prior $p(z)$. We model the distribution of the latent variable $z$ in the variational autoencoder by using a densely connected MoVE layer with $3$ experts. Using multiple experts enables us to capture a richer class of distributions than a single Gaussian distribution could, as is usually the case in VAEs.  We can interpret this as $z$ following a Gaussian Mixture Model, whose components are mutually exclusive and modeled by experts. We integrate the generated data by optimizing a mixture of the loss on the new task data and the loss of the generated data:
\begin{equation}
L(\theta) = \frac{1}{2\vert B_{t}\vert } \sum_{b\in B_t}\ell(b) + \frac{1}{2\vert B_{1:t-1}\vert } \sum_{b\in \vert B_{1:t-1}\vert }\ell(b),
\end{equation}
where $B_t$ is batch of data from the current task, $B_{1:t-1}$ a batch of generated data (instead of stored data from previous tasks), and $\ell(b)$ a loss function on the batch $b$. We were able to improve our results in the supervised settings by incorporating a generative component as a replay mechanism, as we show in Table~\ref{tab:mnistresults}, and in Figure~\ref{tab:cifarresults}.

\subsection{Continual Reinforcement Learning}
\label{sec:crl}
In the continual reinforcement learning (CRL) setting, the agent is tasked with finding an optimal policy in sequentially arriving reinforcement learning problems. To benchmark our method in this setting, we follow the experimental protocol of \cite{ahn2019uncertainty} and use a series of reinforcement learning problems from the PyBullet environments \cite{coumans2021pybullet,benelot2018pybulletgym}. In particular, we use the following: Walker2D, Half Cheetah, Ant, Inverted Double Pendulum, and Hopper. The environments we selected have different states and action dimensions. This implies we can't use a single neural network to model policies and value functions. To remedy this, we pad each state and action with zeros to have equal dimensions. The  Ant environment has the highest dimensionality with a state dimensionality of 28 and an action dimensionality of 8. All others are zero-padded to have this dimensionality. We provide hyper-parameters and other settings in Appendix \ref{app:crlexperiments}.

Our approach to continual reinforcement learning can build upon any deep reinforcement learning algorithm (see Wang et al.\ \cite{wang2020deep} for a review of current algorithms). Here, we chose soft actor-critic (SAC) \cite{haarnoja2018soft}. We extend SAC by implementing all neural networks with MovE layers. When a new task arrives, the old posterior over the expert parameters and the gating posterior become the new priors. After each update step in task $t$, we evaluate the agent in all previous tasks $T_{1:t}$ for three episodes each. We divide the reward achieved during evaluation by the mean reward during training and report the cumulative normalized reward, which gives an upper bound of $t$ in the $t$-th task.

\begin{figure*}[t!]
\begin{minipage}{\textwidth}
\centering
\includegraphics[width=\textwidth, trim={4cm 5cm 5cm 5cm}, clip]{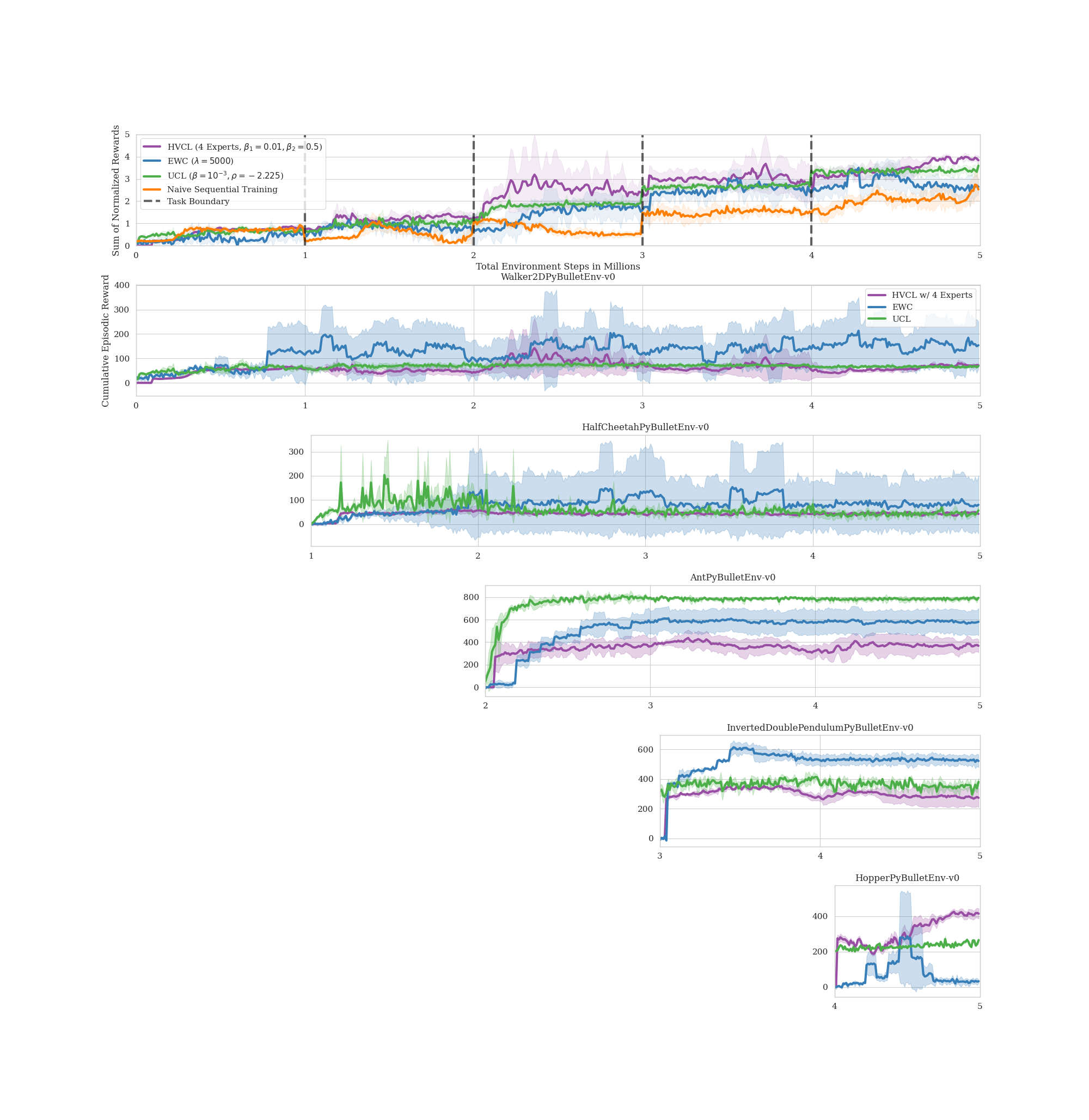}
\end{minipage}
\caption{Continual reinforcement learning. The upper row shows the cumulative normalized rewards over time across the five reinforcement learning tasks. Each vertical dotted line indicates a change in environment after 1 million frames. Performance is computed according to Equation \eqref{eq:crlperformance} after each frame and thus represents the total reward achieved by the agent summed over $t$ tasks (with a maximum normalized reward of one per task). Thus, a value close to $t$ indicates no forgetting, while $1.0$ shows the total forgetting of old tasks. The lower row shows the absolute cumulative episodic reward in each of the five environments given pre-training on the preceding tasks. We compare against EWC \cite{kirkpatrick2017overcoming} and UCL \cite{ahn2019uncertainty}. The lower bound is given by a naively trained dense neural network using SAC \cite{haarnoja2018soft} without CL and without access to old environments.}
\label{fig:crl}
\end{figure*}

We compare our approach against a simple continuously trained SAC implementation with dense neural networks, EWC \cite{kirkpatrick2017overcoming}, and the recently published UCL \cite{ahn2019uncertainty} method. UCL is similar to our approach in that it also employs Bayesian neural networks, but the weight regularization acts on a per-weight basis. Note that, in contrast to our approach, UCL and EWC both require task information to compute task-specific losses. Our results (see Figure~\ref{fig:crl}) show that our approach can sequentially learn new policies while maintaining an acceptable performance on previously seen tasks. We evaluate the methods by computing the following score $J(t)$ for each time step after training on task $t$ is complete:
\begin{equation} 
J(t) = \sum_{k=1}^t \frac{1}{NR_{\mathrm{avg}}(k, \pi_{k})}\sum_{n=1}^N R_n(k, \pi_{t}),
\label{eq:crlperformance}
\end{equation}
where $N$ is the number of episodes to average over, and $R(k, \pi_{k})$ is the cumulative episodic reward in environment $k$ under policy $\pi_{k}$:
\begin{equation} 
R(k, \pi_{k}) = \mathbb{E}_{a_i\thicksim \pi_{k}(s_i), s\thicksim p_k(s_{i+1}\vert a_i, s_i)}\left[r(s_i, a_i)\right],
\end{equation}
where $p_k(s_{i+1}\vert a_i, s_i)$ are the dynamics of environment $k$ and $r(s_i,a_i)$ is the immediate reward for executing action $a_i$ in state $s_i$ at time step $i$.  The policy $\pi_{k}$ refers to the policy trained on tasks up to $k$. The normalization is done by dividing by the average performance over ten episodes of the agent in task $k$, when the agent was trained on that task for the first time. Thus we get a score that measures how well the agent retains its performance on a past environment while learning to solve new problems. As we sum over past environments, an increasing score indicates a successful trade-off between learning and forgetting, while a decreasing or stagnating score indicates forgetting of old tasks.

Our method outperforms UCL \cite{ahn2019uncertainty} and EWC \cite{kirkpatrick2017overcoming}. In this setting naively training the agent sequentially (labeled "Dense") yields poor performance. This behavior indicates the complete forgetting of old policies. The bottom row of Figure \ref{fig:crl}  shows the performance of our method in any particular environment, when pre-training on the preceding environments. It shows that the other methods (UCL and EWC) adapt more successfully to the individual tasks, which is however, coupled with catastrophic forgetting, when switching to the next task. In contrast, our method achieves a better trade-off between learning and forgetting. For this result, we did not optimize hyper-parameters for single task performance, but we simply set the hyper-parameters such that the training performance in each task was comparable in order of magnitude to the other methods. The evaluation across tasks with the same normalized metric shows then the superior ability of HVCL to maintain performance over a sequence of reinforcement learning tasks. Additionally, we note that the variance of the results achieved by our method are lower, suggesting a more stable and reliable training phase.

\subsection{Ablation Studies}
\label{sec:ablation}
To further investigate the methods we propose in this work, we designed a set of ablation experiments. In particular, we aim to demonstrate the importance of each component. To this effect, we run experiments investigating the generator quality in the generative CL setting, study the diversity bonuses in the supervised CL scenario, and take a closer look at the number of experts and the influence of the $\DKL$ weights in the continual reinforcement learning setup.

\subsubsection{Investigating Diversity Bonuses}
\begin{figure*}[t!]
\centering
\includegraphics[width=0.75\textwidth, trim={4cm 3cm 5cm 5cm}, clip]{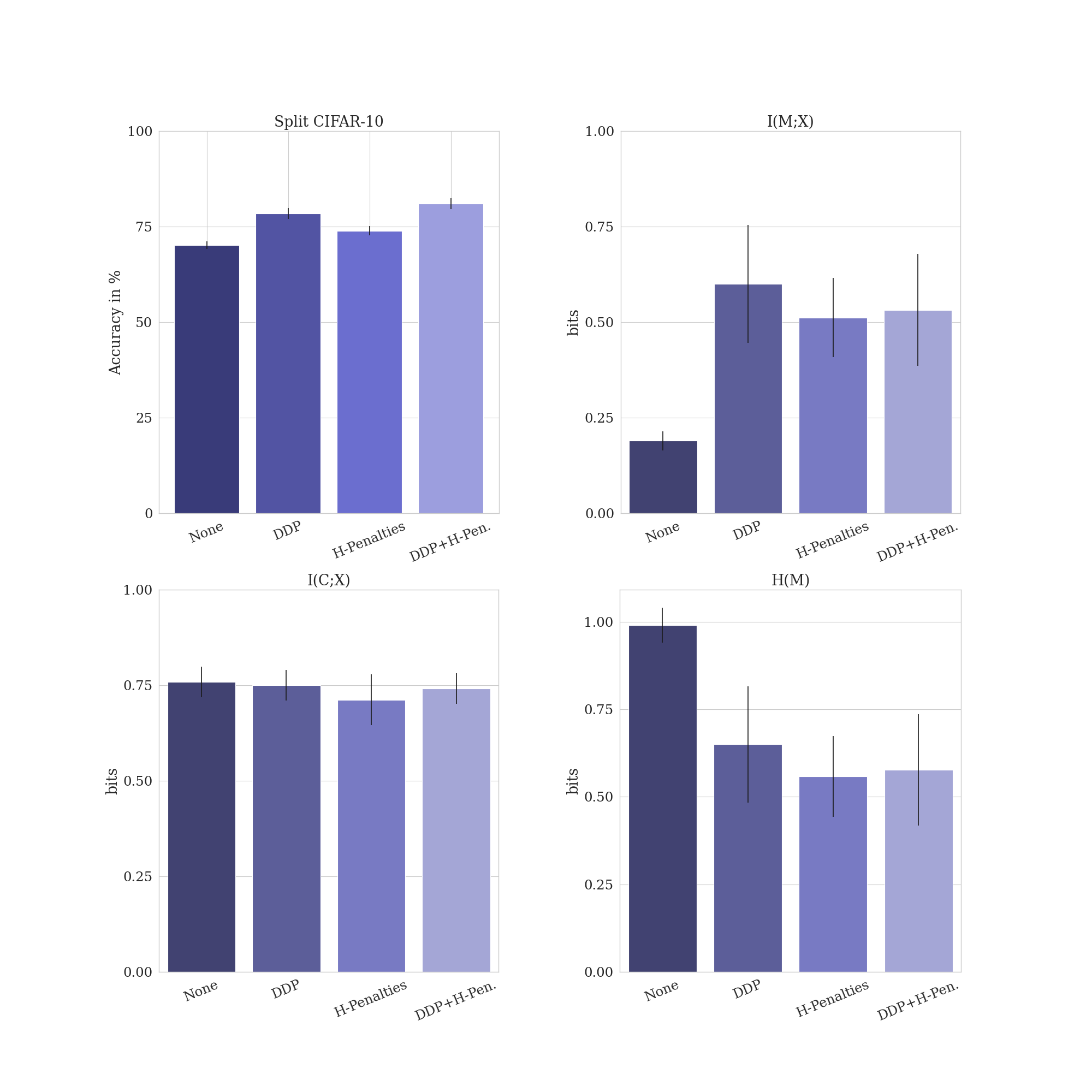}
\caption{Here we evaluate the proposed diversity measures in the split CIFAR-10 benchmark. We averaged every experiment over three random trials. Information-theoretic quantities $I(M;X), I(C;X),$ and $H(X)$ were measured for each layer and averaged.} 
\label{fig:inf_quants}
\end{figure*}

In Section \ref{subsec:diversity} we introduced a diversity objective to stabilize learning in a mixture-of-experts system. Additionally, we argued in favor of an entropy bonus to encourage a selection policy that favors high certainty and sparsity. To investigate the validity of these additions, we run a set of experiments on the Split CIFAR-10 dataset as described in Section \ref{sec:experiments}, but with different bonuses -- see Figure \ref{fig:inf_quants}. In the baseline setup, we used no other objectives as those described by Equation \eqref{eq:hvcl}.

Apart from the classification accuracy, we are interested in three information-theoretic quantities that allow us to investigate the system closer. Firstly, the mutual information between the data generating distribution $p(x)$ and the expert selection $p(m\vert x)$ as measured by $I(M;X)$ indicates how much uncertainty over $m$ the gating unit can reduce on average after observing an input $x$. A higher value means that inputs are differentiated better, which is what we would expect from a more diverse set of experts. $I(M;X)$ is the highest when we use a DPP-based diversity objective ("DDP"), while the entropy of selection policy $H(M)$ is lowest when we use an entropy-based diversity measure ("H-Penalties"), which both show that the objectives we introduced in this study yield the intended results. Combining both the DDP diversity bonus and the entropy penalty on the expert ("DDP+H-Pen.") enforces a trade-off between both objectives and yields the best empirical results.  We average the results of ten random seeds in each setting.

\subsubsection{Generator Quality}
 \begin{figure*}[t!]
\centering
\begin{minipage}{0.66\textwidth}
\includegraphics[width=\textwidth, trim={2.25cm 0.25cm 3cm 1cm}, clip]{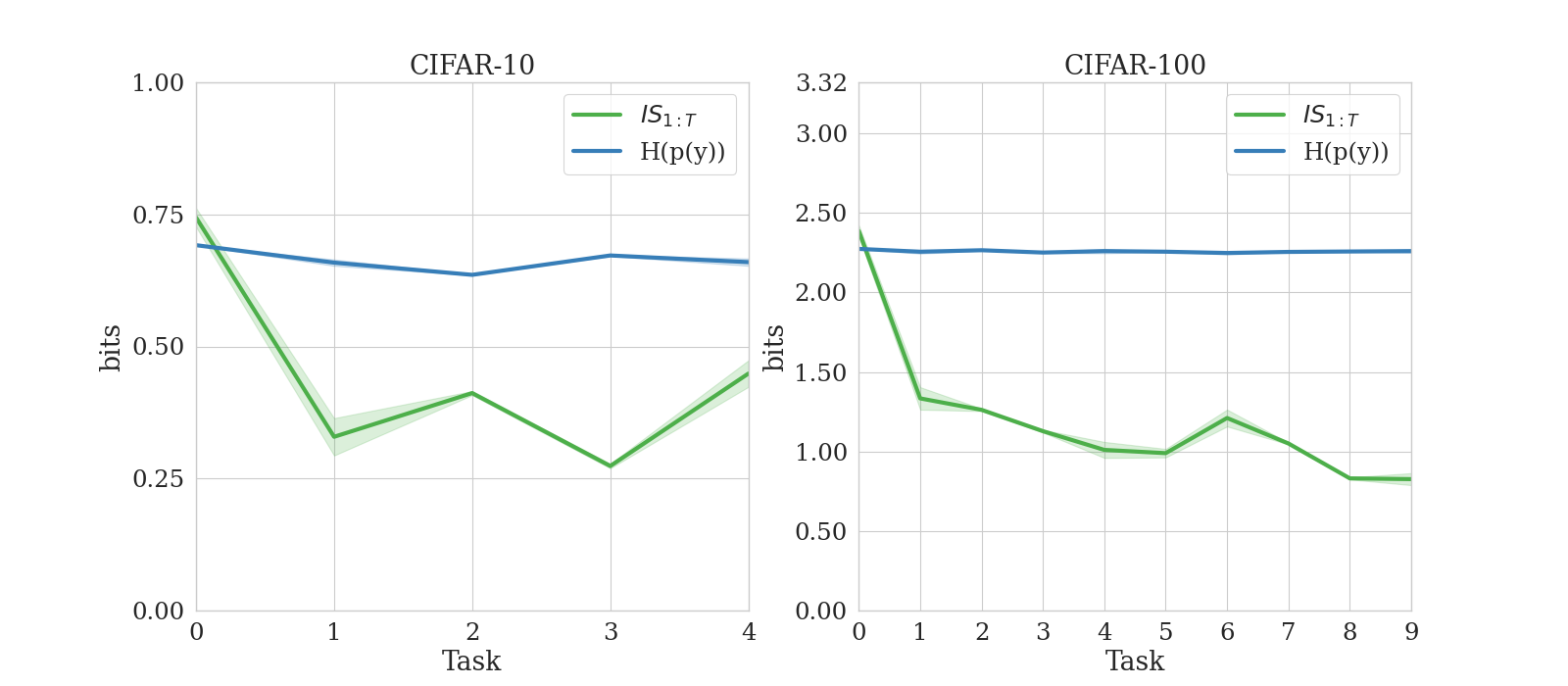}
\end{minipage} 
\begin{minipage}{0.32\textwidth}
\includegraphics[width=\textwidth, trim={0cm 2.15cm 0cm 0cm}, clip]{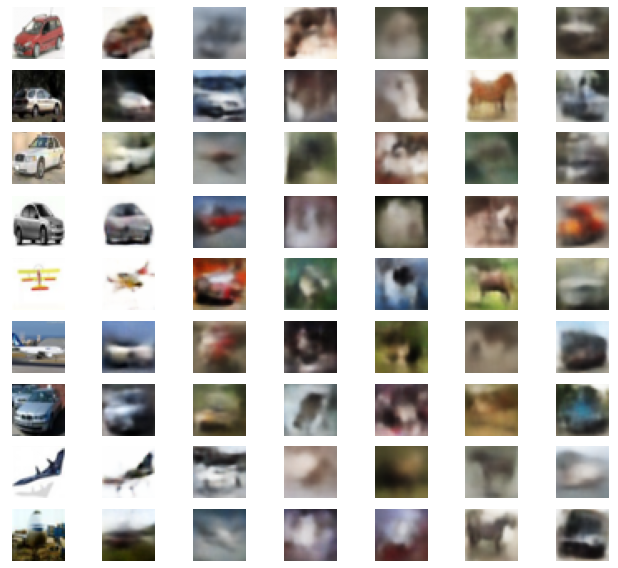}
\end{minipage} 
\caption{Left and middle: Information-theoretic measures for the generator quality. Right: The first two rows show original images and the corresponding reconstructions. The following five rows show samples generated from the VAE after each task (automobiles vs.\ airplanes, birds vs.\ cats, deers vs.\ frogs, dogs vs.\ horses, and ships vs.\ trucks).}
\label{fig:genquality}
\end{figure*}
We introduce a generative approach to continual learning in Section \ref{subsec:vae} by implementing a Variational Auto-encoder using our proposed layer design. This addition improved classification performance and mitigated catastrophic forgetting, as evidenced by the results shown in Figure \ref{tab:cifarresults} and Table \ref{tab:mnistresults}.

By borrowing methods from the generative learning community, we can investigate the performance further. The main focus lies on the quality of the generated images. We can not straightforwardly measure the accuracy, as artificial images lack labels. Thus we first use the trained classifier to obtain labels and compute metrics based on these self-generated labels. We opted for the Inception Score (IS) \cite{salimans2016improved}, as it is widely used in the generative learning community. In this initially proposed formulation, the IS builds on the $\DKL$ between the conditional and the marginal class probabilities as returned by a pre-trained Inception model \cite{szegedy2015going}. To investigate the quality of the generated images concerning the continually trained classifier, we use a different version of the Inception Score, which we defined as
\begin{equation}
IS_{T}(G_{1:T}) = \mathbb{E}_{x\thicksim G_{1:T}}\left[ \DKL\left[p_{1:T}(y\vert x) \vert \vert  p_{1:T}(y)\right] \right],
\end{equation}
where $G_{1:T}$ is the data generator trained on tasks up to $T$, $p_{1:T}(y\vert x)$ the conditional class distribution returned by the classifier trained up to task $T$, and $p_{1:T}(y)$ the marginal class distribution up to Task $T$. Note that, $IS(G_{1:T}) \leq \log_2 N_c$, where $N_c$ is the number of classes. We show $IS_{T}$ and the entropy of $p(y)$ in the split CIFAR-10 and CIFAR-100 setting in Figure~\ref{fig:genquality}. In both cases, it can be seen that the generated pictures retain task-specific information, although there is a notable decrement across tasks.

\subsubsection{Number of Experts}
\label{sec:ablexp}
\begin{figure*}
\includegraphics[width=\textwidth, trim={8.5cm 18cm 8cm 3.5cm}, clip]{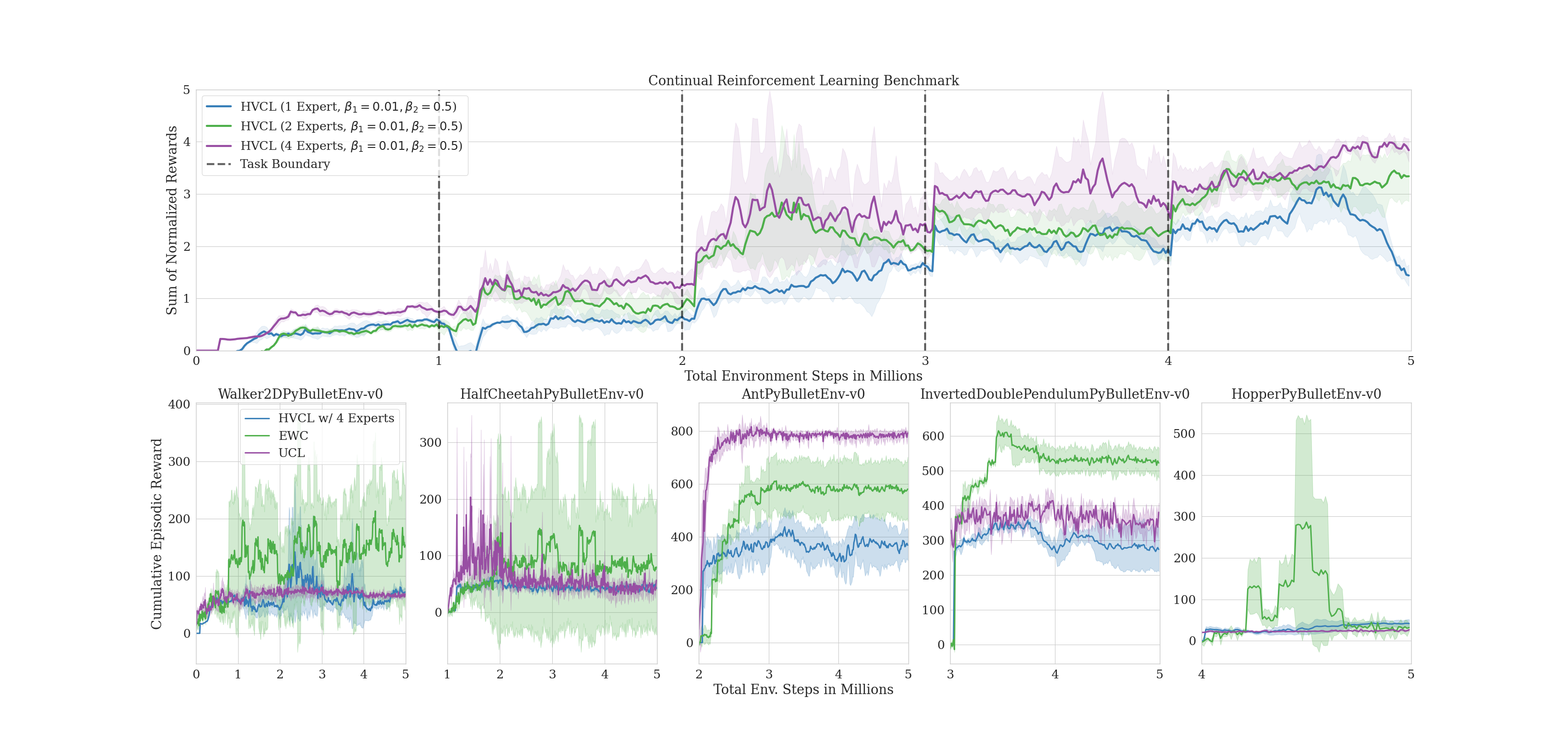}
\caption{Experimental results for systems with 1, 2, and 4 experts. As expected, adding experts mitigates forgetting. Each curve represents three trials in the continual reinforcement learning domain as described in Section \ref{sec:crl}.}
\label{fig:ablexp}
\end{figure*} 
Our method builds on a mixture of experts model and it is thus natural to assume that increasing the number of experts improves performance. Indeed, this is the case as we demonstrate in additional continual reinforcement learning experiments in Figure \ref{fig:ablexp}. As Figure \ref{fig:cl} illustrates, adding experts to layers increases the number of possible information processing paths through the network. Equipped with a diverse and specialized set of parameters, each path can be regarded as a distinct sub-network that learns to solve tasks. 

\subsubsection{$\DKL$ Weights}
\label{sec:abldkl}
As with any hyper-parameter, setting a specific value for $\beta$ has a strong influence on the outcome of the experiments. Setting it too small will lead to the regularization term dominating the loss, and the experts can't learn a new task, as the new parameters remain close to the parameters of the previous task. A high value will drive the penalty term towards zero, which, in turn, will not preserve parameters from old tasks. In principle, there are three ways to choose $\beta$. 

\begin{figure}[t!]
\centering
\includegraphics[width=0.5\textwidth]{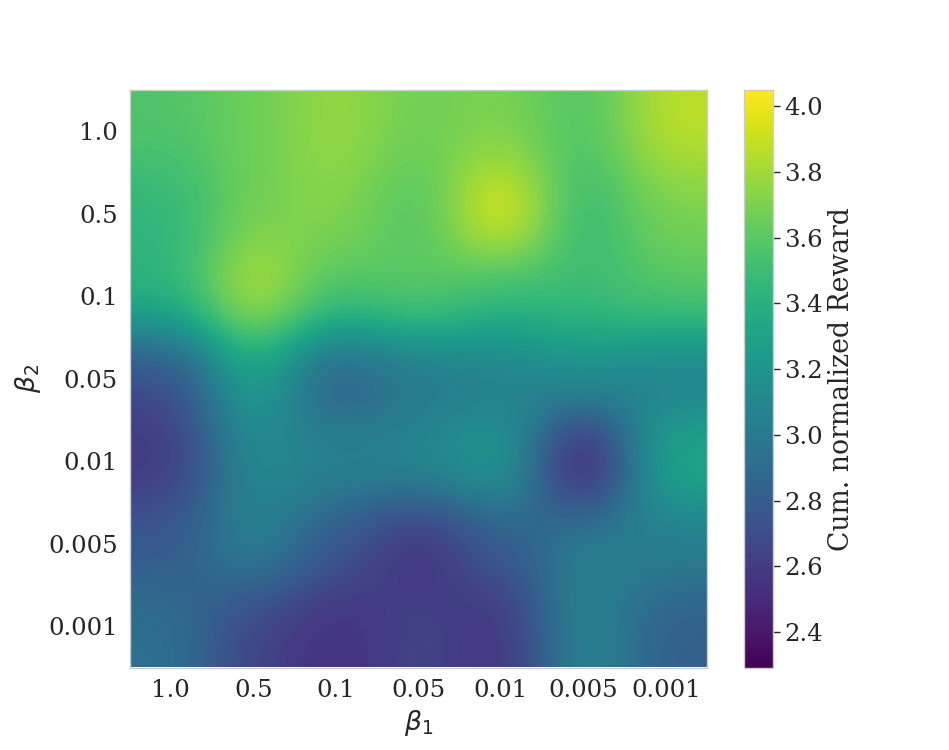}
\caption{In this figure, we show the influence of the $\DKL$ weights $\beta_1$ and $\beta_2$ in the continual reinforcement learning setting. Setting the expert $\DKL$ weight $\beta_2 < 0.1$ results in poor performance, as posteriors deviate too much from their priors.  On the other hand, a lower gating $\DKL$ weight $\beta_1$ allows for a flexible expert allocation and improves performance.}
\label{fig:betaplot}
\end{figure}

First, by setting $\beta$ such that it satisfies an expert information-processing limit. This technique has the advantage that we can interpret this value, e.g.,\ "each expert can process 1.57 bits of information on average, i.e.,\ distinguishing between three options", but shifts the burden from picking $\beta$ to setting a target entropy (see, e.g.,\ Haarnoja et al.\  \cite{haarnoja2018soft} and Grau et al.\ \cite{grau2018soft} for an example of this approach). Second, employing a schedule for $\beta$, as, e.g.,\ proposed by Fu et al.\ \cite{fu2019cyclical}. Last, another option is to run a grid search over a pre-defined range and choose the one that fits best. In our supervised learning experiments, we used a cyclic schedule for $\beta_1$ and $\beta_2$ \cite{fu2019cyclical} while we kept them fixed in the reinforcement learning experiments. To systematically investigate the influence of these parameters, we conducted additional experiments (see Figure \ref{fig:betaplot}).

\section{Discussion}
\label{sec:discussion}
\subsection{Related Work}
\label{sec:related}
The principle we propose in this work falls into a wider class of methods that deal more efficiently with learning and decision-making problems by integrating information-theoretic cost functions. Such information-constrained machine learning methods have enjoyed recent interest in a variety of research fields, such as reinforcement learning \cite{eysenbach2018diversity,ghosh2018divide,leibfried2019mutual,hihn2019information,arumugam2021information}, MCMC optimization \cite{hihn2018bounded,pang2020learning}, meta-learning \cite{rothfuss2018promp,hihn2020hierarchical}, continual learning \cite{nguyen2017variational,ahn2019uncertainty}, and self-supervised learning \cite{thiam2021multi,tsai2021self}. 

The hierarchical structure we employ is a variant of the Mixture of Experts (MoE) model. Jacobs et al.\ \cite{Jacobs1991} introduced MoE as tree-structured models for complex classification and regression problems, where the underlying approach is the divide and conquer paradigm. As in our approach, three main building blocks define MoEs: gates, experts, and a probabilistic weighting to combine expert predictions. Learning proceeds by finding a soft partitioning of the input space and assigning partitions to experts performing well on the partition. MoEs for machine learning have seen a growing interest, with a recent surge stemming from the introduction of the sparsely-gated mixture-of-experts layer \cite{Shazeer2017}. In its initial form, this layer is optimized to divide the inputs equally among experts and then sparsely activate the top-$k$ experts per input. This allowed training systems with billions of total parameters, as only a small subset was active at any given time. In our work, we removed the incentive to equally distribute inputs, as we aim to find specialized experts, which contradicts a balanced load. The computational advantage remains, as we still activate only the top-$1$ expert. 

We extended the sparse MoE layer to continual learning by re-formulating its main principle as a hierarchical extension of variational continual learning \cite{nguyen2017variational}. Our main contribution is removing the need for multi-headed networks by moving the head selection to the gating network. Our method is similar to the approach described in \cite{hihn2020specialization} but differs in two key aspects. Firstly, we provide a more stable learning procedure as our layers can readily offer end-to-end training, which alleviates problems such as expert class imbalance and brittle converging properties reported in the previous study. Secondly, we implement the information-processing constraints on the parameters instead of the output of the experts, thus shifting the information cost from decision-making to learning. A method similar to ours is conditional computing for continual learning \cite{lin2019conditional}. The authors propose to condition the parameters of a neural network on the input samples by learning a (deterministic) function that groups inputs and maps a set of parameters to each group. Our approach differs in two main ways. First, our method can capture uncertainty allowing us to learn stochastic tasks. Second, our design can incorporate up to $2^n$ paths (or groupings) through a neural net with $n$ layers, making it more flexible than learning a mapping function. Another routing approach is routing networks by Collier et al.\ \cite{collier2020routing}. The authors propose to use mixture-of-experts layers that they train with a novel algorithm called co-training. In co-training, an additional data structure keeps track of all experts assigned to a specific task, as these are trained differently than those unassigned so far. In contrast, our method does not require additional training procedures and does not produce any organizational overhead. PathNet \cite{fernando2017pathnet} is a modular neural network architecture where an evolutionary algorithm combines modules (e.g.,\ convolution, max pooling, activation) to solve each task. This approach requires two training procedures: one for the evolutionary algorithm and one to adapt the path modules. Our method allows efficient end-to-end training. Lee et al.\ \cite{Lee2020A} propose a Mixture-of-Experts model for continual learning, in which the number of experts increases dynamically. Their method utilizes Dirichlet-Process-Mixtures \cite{antoniak1974mixtures} to infer the number of experts. The authors argue that since the gating mechanism is itself a classifier, training it in an online fashion would result in catastrophic forgetting. To remedy this, they implement a generative model per expert $m$ to model $p(m\vert x)$ and approximate the output as $p(y\vert x) \approx \sum_m p(y\vert x)p(m\vert x)$. In our work, we have demonstrated that it is possible to implement a gating mechanism based only on the input by coupling it with an information-theoretic objective to prevent catastrophic forgetting.

To stabilize expert training we introduced a diversity objective. Diversity measures have witnessed increasing interest in the reinforcement learning community. The ''diversity is all you need'' (DIAYN) paradigm \cite{eysenbach2018diversity} proposes to formulate an information-processing hierarchy similar to Equation \eqref{eq:par_mutual}, where the information bottleneck on the latent variable discards irrelevant information while an entropy bonus increases diversity. DIAYN acts as intrinsic motivation in environments with no rewards and enables efficient policy learning. Lupu et al.\ \cite{lupu2021trajectory} investigate diversity as a means to create a set of policies in a multi-agent environment. They define diversity between two policies based on a generalization of the Jensen-Shannon divergence and optimize this objective and the main goal defined by the environment simultaneously. Parker et al.\ \cite{parker2020effective} introduce a DDP-based method, that aims to promote diversity in the policy space by defining it as a measure of the different states a policy may reach from a given starting state. Dai et al.\ \cite{dai2021diversity} propose another DDP-based method, where the idea is to augment the sampling process in hindsight experience replay \cite{andrychowicz2017hindsight} (a method for off-policy reinforcement learning) with a diversity bonus. The method we propose to encourage expert diversity differs from previous methods as we define diversity in parameter space instead of the policy outcomes or inputs. In combination with a $\DKL$ penalty, this allows us to optimize for efficient information-processing, which facilities efficient continual learning. Additionally, as we define it on parameters instead of actions, we can apply it straightforwardly to any problem formulation, as our extensive experiments show.

Currently, there are only few methods that perform well in supervised continual learning and continual reinforcement learning \cite[e.g.,][]{ahn2019uncertainty,jung2020continual,cha2020cpr}. These methods require task information, as the either keep a set of separate task-specific heads \cite{ahn2019uncertainty,jung2020continual} or compute task-specific losses \cite{cha2020cpr}. We were able to achieve comparable results to Uncertainty-based continual learning (UCL) \cite{ahn2019uncertainty}. This makes our method one of the first task-agnostic CL approaches to do so.

\subsection{Critical Issues and Future Work}
\label{sec:future}
One drawback of our method is the number of hyper-parameters it introduces. Specifically, the weighting factors for the gating and the expert $\DKL$ constraint, the diversity bonus and its kernel parameters, and the number of experts and the top-$k$ settings. To optimize them in our experiments, we ran a hyperparameter optimization algorithm on a reduced variant of the problem and fine-tuned them on the complete dataset. To further mitigate this problem, we used a scheduling scheme for the $\DKL$ weights, as discussed in more detail in Section \ref{sec:abldkl}.

Moreover, variational inference in neural networks can be computationally expensive \cite{gal2016bayesian,zhang2018noisy,Freitas2000sequential}, as one has to draw samples for each forward pass to approximate gradients. We tackle this issue in three ways. Firstly, we use $D-$dimensional Gaussian mean-field approximate posteriors $q_t(\theta)= \prod^D_{d=1}\mathcal{N}(\theta_t\vert p_{t,d},\sigma^2_{t,d})$ to model distributions. This allows us to compute the $\DKL$ (as well as the diversity measures) in closed form. Secondly, we use the flip-out estimator \cite{wen2018flipout} to approximate the gradients, which is known to have a low variance. In practice, we draw a single sample to approximate the expectation.  Lastly, our top$-k$ sampling allows us to only activate $k$ of the posteriors, which means that the remaining parameters have zero gradient, and accordingly do not have to be updated \cite{lazyadam}. As a consequence, our HVCL approach requires a similar amount of training episodes and computation time as non-probabilistic network representations.

We have observed in simple toy problems that expert allocation and optimization greatly depend on the initialization of the experts. A sufficiently diverse expert initialization yielded better results while requiring fewer iterations. We could not directly transfer this to more complex learning problems because this would introduce computational instabilities during back-propagation, as we lose the variance reducing benefits of state-of-the-art initialization schemes \cite{glorot2010understanding,he2015delving,narkhede2021review}. This is also part of the reason why we chose to implement a diversity objective instead of a novel initializer. We leave this as a topic for future work.

To optimize the diversity measure, we have to compute the determinant of the kernel matrix (see Section \ref{subsec:diversity}) by evaluating the kernel function for each expert pair and finding the derivative of its determinant. While we showed that the Wasserstein-2 distance between Gaussian distributions with diagonal covariances is available in closed form (see Equation \eqref{eq:gaussianwasser}), and that the resulting kernel matrix is symmetric, there is still the issue of the derivative of the determinant. Following Jacobi's formula to do so, requires finding the inverse of the kernel matrix. This operation fails when the kernel matrix is not invertible, which is equivalent to the determinant being zero. This is the case if diversity is also zero, i.e., the expert parameters are pairwise nearly identical. We countered this by setting the kernel width $h$ (see Equation \eqref{eq:w2expkernel}) to a sufficiently large value, which we found by a simple grid search. This problem requires further investigation, as it can impede the complete optimization process.

Our model requires a fixed number of experts. In a more realistic continual learning setting, the number of tasks may grow such that the system may benefit from additional experts. This could be realized by taking a Bayesian non-parametric approach which treats the number of experts as a variable. Dirichlet-Process-Mixtures \cite{antoniak1974mixtures} offer such flexibility, with recent applications to meta-learning \cite{jerfel2019reconciling}, and to continual learning \cite{Lee2020A}.

\section{Conclusion}
\label{sec:conclusion}
We introduced a novel hierarchical approach to task-agnostic continual learning, derived an immediate application, and extensively evaluated this method in supervised continual learning and continual reinforcement learning. The method we introduced builds on a hierarchical Bayesian inference view of continual learning and is a direct extension of Variational Continual Learning (VCL). This adaptive mechanism allowed us to remove the need for extrinsic task-relevant information and to operate in a task-agnostic way. While we removed this limitation, we achieved results competitive to task-aware and to task-agnostic algorithms. These insights allowed us to design a diversity objective that stabilizes learning and further reduces the risk of catastrophic forgetting. In particular, we could show that enforcing expert diversity through additional objectives stabilizes learning and further reduces the risk of catastrophic forgetting. Essentially, catastrophic forgetting is avoided by having multiple experts that are responsible for different tasks and are activated in different contexts and trained only with their assigned data, which avoids weight overwriting and negative interference. Finally, having more diverse experts leads to crisper task partitioning with less interference.

As our method builds on generic utility functions, we can apply it independently of the underlying optimization problem, which makes our method one of the first to do so and achieve competitive results in continual supervised learning benchmarks based on variants of the MNIST, CIFAR-10, and CIFAR-100 datasets. In continual reinforcement learning, we evaluated our method on a series of challenging simulated robotic control tasks. We also demonstrated how our method gives rise to a generative continual learning method. Additionally, we conducted ablation experiments to analyze our approach in a detailed and systematic way. These experiments confirmed that the additional objectives we introduced enhance expert partitioning and enforce a sparse expert selection policy. This leads to specialized and diverse experts, which alleviates catastrophic forgetting. We also investigated the impact of the hyper-parameters our method introduces to highlight how they help to mitigate catastrophic forgetting. In the generative setting, we took a closer look at the performance of the generative model by introducing a continual learning version of the widely used Inception Score. Finally, we showed that increasing the number of experts reduces forgetting in the continual reinforcement learning scenarios.

\section*{Declarations}
\subsection*{Funding}
This work was supported by the European Research Council, grant number ERC-StG-2015-ERC, Project
ID: 678082, ``BRISC: Bounded Rationality in Sensorimotor Coordination".

\subsection*{Declaration of Competing Interest}
All authors certify that they have no affiliations with or involvement in any organization or entity with any financial interest or non-financial interest in the subject matter or materials discussed in this manuscript.

\subsection*{Ethics Approval}
Not applicable.

\subsection*{Code Availability}
Open Source code implementing MoVE Layers is available under \url{https://github.com/hhihn/HVCL/}
\subsection*{Consent to Participate}
Not applicable.

\subsection*{Availability of Data and Material}
Datasets and Environments used in this study: MNIST \url{http://yann.lecun.com/exdb/mnist/}, CIFAR10/1000: \url{https://www.cs.toronto.edu/~kriz/cifar.html}, PyBullet Gymperium: \url{https://github.com/benelot/pybullet-gym}, Tensorflow: \url{https://www.tensorflow.org/}, NumPy \url{https://numpy.org/doc/}, SciPy \url{https://scipy.org/}
\subsection*{Consent for Publication}
Not applicable.

\subsection*{Author Contributions}
\textbf{Heinke Hihn}: Conceptualization, Methodology, Formal analysis, Software,  Visualization, Validation, Investigation, Writing- Original draft preparation - Reviewing and Editing, \textbf{Daniel A. Braun}: Conceptualization, Supervision, Funding acquisition, Writing - Reviewing and Editing,

\newpage
\onecolumn
\bibliographystyle{sn-mathphys} 
\bibliography{merged_bibliography}


\begin{thebibliography}{108}
\ifx \bisbn   \undefined \def \bisbn  #1{ISBN #1}\fi
\ifx \binits  \undefined \def \binits#1{#1}\fi
\ifx \bauthor  \undefined \def \bauthor#1{#1}\fi
\ifx \batitle  \undefined \def \batitle#1{#1}\fi
\ifx \bjtitle  \undefined \def \bjtitle#1{#1}\fi
\ifx \bvolume  \undefined \def \bvolume#1{\textbf{#1}}\fi
\ifx \byear  \undefined \def \byear#1{#1}\fi
\ifx \bissue  \undefined \def \bissue#1{#1}\fi
\ifx \bfpage  \undefined \def \bfpage#1{#1}\fi
\ifx \blpage  \undefined \def \blpage #1{#1}\fi
\ifx \burl  \undefined \def \burl#1{\textsf{#1}}\fi
\ifx \doiurl  \undefined \def \doiurl#1{\url{https://doi.org/#1}}\fi
\ifx \betal  \undefined \def \betal{\textit{et al.}}\fi
\ifx \binstitute  \undefined \def \binstitute#1{#1}\fi
\ifx \binstitutionaled  \undefined \def \binstitutionaled#1{#1}\fi
\ifx \bctitle  \undefined \def \bctitle#1{#1}\fi
\ifx \beditor  \undefined \def \beditor#1{#1}\fi
\ifx \bpublisher  \undefined \def \bpublisher#1{#1}\fi
\ifx \bbtitle  \undefined \def \bbtitle#1{#1}\fi
\ifx \bedition  \undefined \def \bedition#1{#1}\fi
\ifx \bseriesno  \undefined \def \bseriesno#1{#1}\fi
\ifx \blocation  \undefined \def \blocation#1{#1}\fi
\ifx \bsertitle  \undefined \def \bsertitle#1{#1}\fi
\ifx \bsnm \undefined \def \bsnm#1{#1}\fi
\ifx \bsuffix \undefined \def \bsuffix#1{#1}\fi
\ifx \bparticle \undefined \def \bparticle#1{#1}\fi
\ifx \barticle \undefined \def \barticle#1{#1}\fi
\bibcommenthead
\ifx \bconfdate \undefined \def \bconfdate #1{#1}\fi
\ifx \botherref \undefined \def \botherref #1{#1}\fi
\ifx \url \undefined \def \url#1{\textsf{#1}}\fi
\ifx \bchapter \undefined \def \bchapter#1{#1}\fi
\ifx \bbook \undefined \def \bbook#1{#1}\fi
\ifx \bcomment \undefined \def \bcomment#1{#1}\fi
\ifx \oauthor \undefined \def \oauthor#1{#1}\fi
\ifx \citeauthoryear \undefined \def \citeauthoryear#1{#1}\fi
\ifx \endbibitem  \undefined \def \endbibitem {}\fi
\ifx \bconflocation  \undefined \def \bconflocation#1{#1}\fi
\ifx \arxivurl  \undefined \def \arxivurl#1{\textsf{#1}}\fi
\csname PreBibitemsHook\endcsname

\bibitem{mccloskey1989catastrophic}
\begin{bchapter}
\bauthor{\bsnm{McCloskey}, \binits{M.}},
\bauthor{\bsnm{Cohen}, \binits{N.J.}}:
\bctitle{Catastrophic interference in connectionist networks: The sequential
  learning problem}.
In: \bbtitle{Psychology of Learning and Motivation}
vol. \bseriesno{24},
pp. \bfpage{109}--\blpage{165}.
\bpublisher{Elsevier}, \blocation{???}
(\byear{1989}).
\doiurl{10.1016/S0079-7421(08)60536-8}
\end{bchapter}
\endbibitem

\bibitem{thrun1998lifelong}
\begin{bchapter}
\bauthor{\bsnm{Thrun}, \binits{S.}}:
\bctitle{Lifelong learning algorithms}.
In: \bbtitle{Learning to Learn},
pp. \bfpage{181}--\blpage{209}.
\bpublisher{Springer},
\blocation{Heidelberg}
(\byear{1998})
\end{bchapter}
\endbibitem

\bibitem{shin2017continual}
\begin{bchapter}
\bauthor{\bsnm{Shin}, \binits{H.}},
\bauthor{\bsnm{Lee}, \binits{J.K.}},
\bauthor{\bsnm{Kim}, \binits{J.}},
\bauthor{\bsnm{Kim}, \binits{J.}}:
\bctitle{Continual learning with deep generative replay}.
In: \bbtitle{Advances in Neural Information Processing Systems},
pp. \bfpage{2990}--\blpage{2999}
(\byear{2017})
\end{bchapter}
\endbibitem

\bibitem{rebuffi2017icarl}
\begin{bchapter}
\bauthor{\bsnm{Rebuffi}, \binits{S.-A.}},
\bauthor{\bsnm{Kolesnikov}, \binits{A.}},
\bauthor{\bsnm{Sperl}, \binits{G.}},
\bauthor{\bsnm{Lampert}, \binits{C.H.}}:
\bctitle{icarl: Incremental classifier and representation learning}.
In: \bbtitle{Proceedings of the IEEE Conference on Computer Vision and Pattern
  Recognition},
pp. \bfpage{2001}--\blpage{2010}
(\byear{2017})
\end{bchapter}
\endbibitem

\bibitem{wilson1994reactivation}
\begin{barticle}
\bauthor{\bsnm{Wilson}, \binits{M.A.}},
\bauthor{\bsnm{McNaughton}, \binits{B.L.}}:
\batitle{Reactivation of hippocampal ensemble memories during sleep}.
\bjtitle{Science}
\bvolume{265}(\bissue{5172}),
\bfpage{676}--\blpage{679}
(\byear{1994})
\end{barticle}
\endbibitem

\bibitem{rasch2007maintaining}
\begin{barticle}
\bauthor{\bsnm{Rasch}, \binits{B.}},
\bauthor{\bsnm{Born}, \binits{J.}}:
\batitle{Maintaining memories by reactivation}.
\bjtitle{Current opinion in neurobiology}
\bvolume{17}(\bissue{6}),
\bfpage{698}--\blpage{703}
(\byear{2007})
\end{barticle}
\endbibitem

\bibitem{van2016hippocampal}
\begin{barticle}
\bauthor{\bparticle{van~de} \bsnm{Ven}, \binits{G.M.}},
\bauthor{\bsnm{Trouche}, \binits{S.}},
\bauthor{\bsnm{McNamara}, \binits{C.G.}},
\bauthor{\bsnm{Allen}, \binits{K.}},
\bauthor{\bsnm{Dupret}, \binits{D.}}:
\batitle{Hippocampal offline reactivation consolidates recently formed cell
  assembly patterns during sharp wave-ripples}.
\bjtitle{Neuron}
\bvolume{92}(\bissue{5}),
\bfpage{968}--\blpage{974}
(\byear{2016})
\end{barticle}
\endbibitem

\bibitem{kirkpatrick2017overcoming}
\begin{barticle}
\bauthor{\bsnm{Kirkpatrick}, \binits{J.}},
\bauthor{\bsnm{Pascanu}, \binits{R.}},
\bauthor{\bsnm{Rabinowitz}, \binits{N.}},
\bauthor{\bsnm{Veness}, \binits{J.}},
\bauthor{\bsnm{Desjardins}, \binits{G.}},
\bauthor{\bsnm{Rusu}, \binits{A.A.}},
\bauthor{\bsnm{Milan}, \binits{K.}},
\bauthor{\bsnm{Quan}, \binits{J.}},
\bauthor{\bsnm{Ramalho}, \binits{T.}},
\bauthor{\bsnm{Grabska-Barwinska}, \binits{A.}}, \betal:
\batitle{Overcoming catastrophic forgetting in neural networks}.
\bjtitle{Proceedings of the national academy of sciences}
\bvolume{114}(\bissue{13}),
\bfpage{3521}--\blpage{3526}
(\byear{2017})
\end{barticle}
\endbibitem

\bibitem{zenke2017continual}
\begin{barticle}
\bauthor{\bsnm{Zenke}, \binits{F.}},
\bauthor{\bsnm{Poole}, \binits{B.}},
\bauthor{\bsnm{Ganguli}, \binits{S.}}:
\batitle{Continual learning through synaptic intelligence}.
\bjtitle{Proceedings of machine learning research}
\bvolume{70},
\bfpage{3987}
(\byear{2017})
\end{barticle}
\endbibitem

\bibitem{ahn2019uncertainty}
\begin{bchapter}
\bauthor{\bsnm{Ahn}, \binits{H.}},
\bauthor{\bsnm{Cha}, \binits{S.}},
\bauthor{\bsnm{Lee}, \binits{D.}},
\bauthor{\bsnm{Moon}, \binits{T.}}:
\bctitle{Uncertainty-based continual learning with adaptive regularization}.
In: \bbtitle{Proceedings of the 33rd International Conference on Neural
  Information Processing Systems},
pp. \bfpage{4392}--\blpage{4402}
(\byear{2019})
\end{bchapter}
\endbibitem

\bibitem{benavides2020towards}
\begin{barticle}
\bauthor{\bsnm{Benavides-Prado}, \binits{D.}},
\bauthor{\bsnm{Koh}, \binits{Y.S.}},
\bauthor{\bsnm{Riddle}, \binits{P.}}:
\batitle{Towards knowledgeable supervised lifelong learning systems}.
\bjtitle{Journal of Artificial Intelligence Research}
\bvolume{68},
\bfpage{159}--\blpage{224}
(\byear{2020})
\end{barticle}
\endbibitem

\bibitem{han2021continual}
\begin{bchapter}
\bauthor{\bsnm{Han}, \binits{X.}},
\bauthor{\bsnm{Guo}, \binits{Y.}}:
\bctitle{Continual learning with dual regularizations}.
In: \bbtitle{Joint European Conference on Machine Learning and Knowledge
  Discovery in Databases},
pp. \bfpage{619}--\blpage{634}
(\byear{2021}).
\bcomment{Springer}
\end{bchapter}
\endbibitem

\bibitem{li2021lifelong}
\begin{bchapter}
\bauthor{\bsnm{Li}, \binits{H.}},
\bauthor{\bsnm{Krishnan}, \binits{A.}},
\bauthor{\bsnm{Wu}, \binits{J.}},
\bauthor{\bsnm{Kolouri}, \binits{S.}},
\bauthor{\bsnm{Pilly}, \binits{P.K.}},
\bauthor{\bsnm{Braverman}, \binits{V.}}:
\bctitle{Lifelong learning with sketched structural regularization}.
In: \bbtitle{Asian Conference on Machine Learning},
pp. \bfpage{985}--\blpage{1000}
(\byear{2021}).
\bcomment{PMLR}
\end{bchapter}
\endbibitem

\bibitem{cha2020cpr}
\begin{bchapter}
\bauthor{\bsnm{Cha}, \binits{S.}},
\bauthor{\bsnm{Hsu}, \binits{H.}},
\bauthor{\bsnm{Hwang}, \binits{T.}},
\bauthor{\bsnm{Calmon}, \binits{F.}},
\bauthor{\bsnm{Moon}, \binits{T.}}:
\bctitle{Cpr: Classifier-projection regularization for continual learning}.
In: \bbtitle{International Conference on Learning Representations}
(\byear{2020})
\end{bchapter}
\endbibitem

\bibitem{ostapenko2019learning}
\begin{bchapter}
\bauthor{\bsnm{Ostapenko}, \binits{O.}},
\bauthor{\bsnm{Puscas}, \binits{M.}},
\bauthor{\bsnm{Klein}, \binits{T.}},
\bauthor{\bsnm{Jahnichen}, \binits{P.}},
\bauthor{\bsnm{Nabi}, \binits{M.}}:
\bctitle{Learning to remember: A synaptic plasticity driven framework for
  continual learning}.
In: \bbtitle{Proceedings of the IEEE/CVF Conference on Computer Vision and
  Pattern Recognition},
pp. \bfpage{11321}--\blpage{11329}
(\byear{2019})
\end{bchapter}
\endbibitem

\bibitem{lin2019conditional}
\begin{bchapter}
\bauthor{\bsnm{Lin}, \binits{M.}},
\bauthor{\bsnm{Fu}, \binits{J.}},
\bauthor{\bsnm{Bengio}, \binits{Y.}}:
\bctitle{Conditional computation for continual learning}.
In: \bbtitle{NeurIPS 2018 Continual Learning Workshop}
(\byear{2019})
\end{bchapter}
\endbibitem

\bibitem{fernando2017pathnet}
\begin{bchapter}
\bauthor{\bsnm{Fernando}, \binits{C.}},
\bauthor{\bsnm{Banarse}, \binits{D.}},
\bauthor{\bsnm{Blundell}, \binits{C.}},
\bauthor{\bsnm{Zwols}, \binits{Y.}},
\bauthor{\bsnm{Ha}, \binits{D.}},
\bauthor{\bsnm{Rusu}, \binits{A.A.}},
\bauthor{\bsnm{Pritzel}, \binits{A.}},
\bauthor{\bsnm{Wierstra}, \binits{D.}}:
\bctitle{Pathnet: Evolution channels gradient descent in super neural
  networks}.
In: \bbtitle{arXiv Preprint arXiv:1701.08734}
(\byear{2017})
\end{bchapter}
\endbibitem

\bibitem{rusu2016progressive}
\begin{bchapter}
\bauthor{\bsnm{Rusu}, \binits{A.A.}},
\bauthor{\bsnm{Rabinowitz}, \binits{N.C.}},
\bauthor{\bsnm{Desjardins}, \binits{G.}},
\bauthor{\bsnm{Soyer}, \binits{H.}},
\bauthor{\bsnm{Kirkpatrick}, \binits{J.}},
\bauthor{\bsnm{Kavukcuoglu}, \binits{K.}},
\bauthor{\bsnm{Pascanu}, \binits{R.}},
\bauthor{\bsnm{Hadsell}, \binits{R.}}:
\bctitle{Progressive neural networks}.
In: \bbtitle{NIPS Deep Learning Symposium}
(\byear{2016})
\end{bchapter}
\endbibitem

\bibitem{yoon2018lifelong}
\begin{bchapter}
\bauthor{\bsnm{Yoon}, \binits{J.}},
\bauthor{\bsnm{Yang}, \binits{E.}},
\bauthor{\bsnm{Lee}, \binits{J.}},
\bauthor{\bsnm{Hwang}, \binits{S.J.}}:
\bctitle{Lifelong learning with dynamically expandable networks}.
In: \bbtitle{6th International Conference on Learning Representations, ICLR
  2018}
(\byear{2018}).
\bcomment{International Conference on Learning Representations, ICLR}
\end{bchapter}
\endbibitem

\bibitem{golkar2019continual}
\begin{bchapter}
\bauthor{\bsnm{Golkar}, \binits{S.}},
\bauthor{\bsnm{Kagan}, \binits{M.}},
\bauthor{\bsnm{Cho}, \binits{K.}}:
\bctitle{Continual learning via neural pruning}.
In: \bbtitle{NeurIPS 2019 Workshop Neuro AI}
(\byear{2019})
\end{bchapter}
\endbibitem

\bibitem{zacarias2018sena}
\begin{bchapter}
\bauthor{\bsnm{Zacarias}, \binits{A.}},
\bauthor{\bsnm{Alexandre}, \binits{L.A.}}:
\bctitle{Sena-cnn: overcoming catastrophic forgetting in convolutional neural
  networks by selective network augmentation}.
In: \bbtitle{IAPR Workshop on Artificial Neural Networks in Pattern
  Recognition},
pp. \bfpage{102}--\blpage{112}
(\byear{2018}).
\bcomment{Springer}
\end{bchapter}
\endbibitem

\bibitem{collier2020routing}
\begin{bchapter}
\bauthor{\bsnm{Collier}, \binits{M.}},
\bauthor{\bsnm{Kokiopoulou}, \binits{E.}},
\bauthor{\bsnm{Gesmundo}, \binits{A.}},
\bauthor{\bsnm{Berent}, \binits{J.}}:
\bctitle{Routing networks with co-training for continual learning}.
In: \bbtitle{ICML 2020 Workshop on Continual Learning}
(\byear{2020})
\end{bchapter}
\endbibitem

\bibitem{zhai2019lifelong}
\begin{bchapter}
\bauthor{\bsnm{Zhai}, \binits{M.}},
\bauthor{\bsnm{Chen}, \binits{L.}},
\bauthor{\bsnm{Tung}, \binits{F.}},
\bauthor{\bsnm{He}, \binits{J.}},
\bauthor{\bsnm{Nawhal}, \binits{M.}},
\bauthor{\bsnm{Mori}, \binits{G.}}:
\bctitle{Lifelong gan: Continual learning for conditional image generation}.
In: \bbtitle{Proceedings of the IEEE/CVF International Conference on Computer
  Vision},
pp. \bfpage{2759}--\blpage{2768}
(\byear{2019})
\end{bchapter}
\endbibitem

\bibitem{liu2020mnemonics}
\begin{bchapter}
\bauthor{\bsnm{Liu}, \binits{Y.}},
\bauthor{\bsnm{Su}, \binits{Y.}},
\bauthor{\bsnm{Liu}, \binits{A.-A.}},
\bauthor{\bsnm{Schiele}, \binits{B.}},
\bauthor{\bsnm{Sun}, \binits{Q.}}:
\bctitle{Mnemonics training: Multi-class incremental learning without
  forgetting}.
In: \bbtitle{Proceedings of the IEEE/CVF Conference on Computer Vision and
  Pattern Recognition},
pp. \bfpage{12245}--\blpage{12254}
(\byear{2020})
\end{bchapter}
\endbibitem

\bibitem{Lee2020A}
\begin{bchapter}
\bauthor{\bsnm{Lee}, \binits{S.}},
\bauthor{\bsnm{Ha}, \binits{J.}},
\bauthor{\bsnm{Zhang}, \binits{D.}},
\bauthor{\bsnm{Kim}, \binits{G.}}:
\bctitle{A neural dirichlet process mixture model for task-free continual
  learning}.
In: \bbtitle{International Conference on Learning Representations}
(\byear{2020}).
\burl{https://openreview.net/forum?id=SJxSOJStPr}
\end{bchapter}
\endbibitem

\bibitem{zeng2019continual}
\begin{barticle}
\bauthor{\bsnm{Zeng}, \binits{G.}},
\bauthor{\bsnm{Chen}, \binits{Y.}},
\bauthor{\bsnm{Cui}, \binits{B.}},
\bauthor{\bsnm{Yu}, \binits{S.}}:
\batitle{Continual learning of context-dependent processing in neural
  networks}.
\bjtitle{Nature Machine Intelligence}
\bvolume{1}(\bissue{8}),
\bfpage{364}--\blpage{372}
(\byear{2019})
\end{barticle}
\endbibitem

\bibitem{wang2021training}
\begin{bchapter}
\bauthor{\bsnm{Wang}, \binits{S.}},
\bauthor{\bsnm{Li}, \binits{X.}},
\bauthor{\bsnm{Sun}, \binits{J.}},
\bauthor{\bsnm{Xu}, \binits{Z.}}:
\bctitle{Training networks in null space of feature covariance for continual
  learning}.
In: \bbtitle{Proceedings of the IEEE/CVF Conference on Computer Vision and
  Pattern Recognition},
pp. \bfpage{184}--\blpage{193}
(\byear{2021})
\end{bchapter}
\endbibitem

\bibitem{biesialska2020continual}
\begin{bchapter}
\bauthor{\bsnm{Biesialska}, \binits{M.}},
\bauthor{\bsnm{Biesialska}, \binits{K.}},
\bauthor{\bsnm{Costa-juss{\`a}}, \binits{M.R.}}:
\bctitle{Continual lifelong learning in natural language processing: A survey}.
In: \bbtitle{Proceedings of the 28th International Conference on Computational
  Linguistics},
pp. \bfpage{6523}--\blpage{6541}
(\byear{2020})
\end{bchapter}
\endbibitem

\bibitem{de2021continual}
\begin{barticle}
\bauthor{\bsnm{De~Lange}, \binits{M.}},
\bauthor{\bsnm{Aljundi}, \binits{R.}},
\bauthor{\bsnm{Masana}, \binits{M.}},
\bauthor{\bsnm{Parisot}, \binits{S.}},
\bauthor{\bsnm{Jia}, \binits{X.}},
\bauthor{\bsnm{Leonardis}, \binits{A.}},
\bauthor{\bsnm{Slabaugh}, \binits{G.}},
\bauthor{\bsnm{Tuytelaars}, \binits{T.}}:
\batitle{A continual learning survey: Defying forgetting in classification
  tasks}.
\bjtitle{IEEE transactions on pattern analysis and machine intelligence}
\bvolume{44}(\bissue{7}),
\bfpage{3366}--\blpage{3385}
(\byear{2021})
\end{barticle}
\endbibitem

\bibitem{vijayan2021continual}
\begin{bchapter}
\bauthor{\bsnm{Vijayan}, \binits{M.}},
\bauthor{\bsnm{Sridhar}, \binits{S.S.}}:
\bctitle{Continual learning for classification problems: A survey}.
In: \beditor{\bsnm{Krishnamurthy}, \binits{V.}},
\beditor{\bsnm{Jaganathan}, \binits{S.}},
\beditor{\bsnm{Rajaram}, \binits{K.}},
\beditor{\bsnm{Shunmuganathan}, \binits{S.}} (eds.)
\bbtitle{Computational Intelligence in Data Science},
pp. \bfpage{156}--\blpage{166}.
\bpublisher{Springer},
\blocation{Cham}
(\byear{2021})
\end{bchapter}
\endbibitem

\bibitem{parisi2019continual}
\begin{barticle}
\bauthor{\bsnm{Parisi}, \binits{G.I.}},
\bauthor{\bsnm{Kemker}, \binits{R.}},
\bauthor{\bsnm{Part}, \binits{J.L.}},
\bauthor{\bsnm{Kanan}, \binits{C.}},
\bauthor{\bsnm{Wermter}, \binits{S.}}:
\batitle{Continual lifelong learning with neural networks: A review}.
\bjtitle{Neural Networks}
\bvolume{113},
\bfpage{54}--\blpage{71}
(\byear{2019})
\end{barticle}
\endbibitem

\bibitem{nguyen2017variational}
\begin{bchapter}
\bauthor{\bsnm{Nguyen}, \binits{C.V.}},
\bauthor{\bsnm{Li}, \binits{Y.}},
\bauthor{\bsnm{Bui}, \binits{T.D.}},
\bauthor{\bsnm{Turner}, \binits{R.E.}}:
\bctitle{Variational continual learning}.
In: \bbtitle{Proceedings of the International Conference on Representation
  Learning}
(\byear{2017})
\end{bchapter}
\endbibitem

\bibitem{li2017learning}
\begin{barticle}
\bauthor{\bsnm{Li}, \binits{Z.}},
\bauthor{\bsnm{Hoiem}, \binits{D.}}:
\batitle{Learning without forgetting}.
\bjtitle{IEEE transactions on pattern analysis and machine intelligence}
\bvolume{40}(\bissue{12}),
\bfpage{2935}--\blpage{2947}
(\byear{2017}).
\doiurl{10.1109/TPAMI.2017.2773081}
\end{barticle}
\endbibitem

\bibitem{rao2019continual}
\begin{botherref}
\oauthor{\bsnm{Rao}, \binits{D.}},
\oauthor{\bsnm{Visin}, \binits{F.}},
\oauthor{\bsnm{Rusu}, \binits{A.}},
\oauthor{\bsnm{Pascanu}, \binits{R.}},
\oauthor{\bsnm{Teh}, \binits{Y.W.}},
\oauthor{\bsnm{Hadsell}, \binits{R.}}:
Continual unsupervised representation learning.
Advances in Neural Information Processing Systems
\textbf{32}
(2019)
\end{botherref}
\endbibitem

\bibitem{sokar2021self}
\begin{bchapter}
\bauthor{\bsnm{Sokar}, \binits{G.}},
\bauthor{\bsnm{Mocanu}, \binits{D.C.}},
\bauthor{\bsnm{Pechenizkiy}, \binits{M.}}:
\bctitle{Self-attention meta-learner for continual learning}.
In: \bbtitle{Proceedings of the 20th International Conference on Autonomous
  Agents and MultiAgent Systems},
pp. \bfpage{1658}--\blpage{1660}
(\byear{2021})
\end{bchapter}
\endbibitem

\bibitem{han2021contrastive}
\begin{botherref}
\oauthor{\bsnm{Han}, \binits{X.}},
\oauthor{\bsnm{Guo}, \binits{Y.}}:
Contrastive continual learning with feature propagation.
arXiv preprint arXiv:2112.01713
(2021)
\end{botherref}
\endbibitem

\bibitem{chaudhry2021using}
\begin{bchapter}
\bauthor{\bsnm{Chaudhry}, \binits{A.}},
\bauthor{\bsnm{Gordo}, \binits{A.}},
\bauthor{\bsnm{Dokania}, \binits{P.}},
\bauthor{\bsnm{Torr}, \binits{P.}},
\bauthor{\bsnm{Lopez-Paz}, \binits{D.}}:
\bctitle{Using hindsight to anchor past knowledge in continual learning}.
In: \bbtitle{Proceedings of the AAAI Conference on Artificial Intelligence},
vol. \bseriesno{35},
pp. \bfpage{6993}--\blpage{7001}
(\byear{2021})
\end{bchapter}
\endbibitem

\bibitem{chaudhry2018efficient}
\begin{bchapter}
\bauthor{\bsnm{Chaudhry}, \binits{A.}},
\bauthor{\bsnm{Ranzato}, \binits{M.}},
\bauthor{\bsnm{Rohrbach}, \binits{M.}},
\bauthor{\bsnm{Elhoseiny}, \binits{M.}}:
\bctitle{Efficient lifelong learning with a-gem}.
In: \bbtitle{International Conference on Learning Representations}
(\byear{2018})
\end{bchapter}
\endbibitem

\bibitem{khatib2019strategies}
\begin{bchapter}
\bauthor{\bsnm{El~Khatib}, \binits{A.}},
\bauthor{\bsnm{Karray}, \binits{F.}}:
\bctitle{Strategies for improving single-head continual learning performance}.
In: \beditor{\bsnm{Karray}, \binits{F.}},
\beditor{\bsnm{Campilho}, \binits{A.}},
\beditor{\bsnm{Yu}, \binits{A.}} (eds.)
\bbtitle{Image Analysis and Recognition},
pp. \bfpage{452}--\blpage{460}.
\bpublisher{Springer},
\blocation{Cham}
(\byear{2019})
\end{bchapter}
\endbibitem

\bibitem{hihn2020specialization}
\begin{barticle}
\bauthor{\bsnm{Hihn}, \binits{H.}},
\bauthor{\bsnm{Braun}, \binits{D.A.}}:
\batitle{Specialization in hierarchical learning systems}.
\bjtitle{Neural Processing Letters}
\bvolume{52}(\bissue{3}),
\bfpage{2319}--\blpage{2352}
(\byear{2020})
\end{barticle}
\endbibitem

\bibitem{yao2019hierarchically}
\begin{bchapter}
\bauthor{\bsnm{Yao}, \binits{H.}},
\bauthor{\bsnm{Wei}, \binits{Y.}},
\bauthor{\bsnm{Huang}, \binits{J.}},
\bauthor{\bsnm{Li}, \binits{Z.}}:
\bctitle{Hierarchically structured meta-learning}.
In: \bbtitle{Proceedings of the International Conference on Machine Learning},
pp. \bfpage{7045}--\blpage{7054}
(\byear{2019})
\end{bchapter}
\endbibitem

\bibitem{chaudhry2018riemannian}
\begin{bchapter}
\bauthor{\bsnm{Chaudhry}, \binits{A.}},
\bauthor{\bsnm{Dokania}, \binits{P.K.}},
\bauthor{\bsnm{Ajanthan}, \binits{T.}},
\bauthor{\bsnm{Torr}, \binits{P.H.}}:
\bctitle{Riemannian walk for incremental learning: Understanding forgetting and
  intransigence}.
In: \bbtitle{Proceedings of the European Conference on Computer Vision (ECCV)},
pp. \bfpage{532}--\blpage{547}
(\byear{2018})
\end{bchapter}
\endbibitem

\bibitem{Shazeer2017}
\begin{bchapter}
\bauthor{\bsnm{Shazeer}, \binits{N.}},
\bauthor{\bsnm{Mirhoseini}, \binits{A.}},
\bauthor{\bsnm{Maziarz}, \binits{K.}},
\bauthor{\bsnm{Davis}, \binits{A.}},
\bauthor{\bsnm{Le}, \binits{Q.}},
\bauthor{\bsnm{Hinton}, \binits{G.}},
\bauthor{\bsnm{Dean}, \binits{J.}}:
\bctitle{Outrageously large neural networks: The sparsely-gated
  mixture-of-experts layer}.
In: \bbtitle{Proceedings of the International Conference on Learning
  Representations (ICLR)}
(\byear{2017})
\end{bchapter}
\endbibitem

\bibitem{Kuncheva2004}
\begin{bbook}
\bauthor{\bsnm{Kuncheva}, \binits{L.I.}}:
\bbtitle{Combining Pattern Classifiers: Methods and Algorithms}.
\bpublisher{John Wiley \& Sons},
\blocation{Hoboken, NJ}
(\byear{2004})
\end{bbook}
\endbibitem

\bibitem{kuncheva2003measures}
\begin{barticle}
\bauthor{\bsnm{Kuncheva}, \binits{L.I.}},
\bauthor{\bsnm{Whitaker}, \binits{C.J.}}:
\batitle{Measures of diversity in classifier ensembles and their relationship
  with the ensemble accuracy}.
\bjtitle{Machine learning}
\bvolume{51}(\bissue{2}),
\bfpage{181}--\blpage{207}
(\byear{2003})
\end{barticle}
\endbibitem

\bibitem{bian2021when}
\begin{botherref}
\oauthor{\bsnm{Bian}, \binits{Y.}},
\oauthor{\bsnm{Chen}, \binits{H.}}:
When does diversity help generalization in classification ensembles.
IEEE Transactions on Cybernetics
(2021)
\end{botherref}
\endbibitem

\bibitem{kulesza2012determinantal}
\begin{barticle}
\bauthor{\bsnm{Kulesza}, \binits{A.}},
\bauthor{\bsnm{Taskar}, \binits{B.}}, \betal:
\batitle{Determinantal point processes for machine learning}.
\bjtitle{Foundations and Trends{\textregistered} in Machine Learning}
\bvolume{5}(\bissue{2--3}),
\bfpage{123}--\blpage{286}
(\byear{2012})
\end{barticle}
\endbibitem

\bibitem{Jacobs1991}
\begin{barticle}
\bauthor{\bsnm{Jacobs}, \binits{R.A.}},
\bauthor{\bsnm{Jordan}, \binits{M.I.}},
\bauthor{\bsnm{Nowlan}, \binits{S.J.}},
\bauthor{\bsnm{Hinton}, \binits{G.E.}}:
\batitle{Adaptive mixtures of local experts}.
\bjtitle{Neural computation}
\bvolume{3}(\bissue{1}),
\bfpage{79}--\blpage{87}
(\byear{1991})
\end{barticle}
\endbibitem

\bibitem{bang2021rainbow}
\begin{bchapter}
\bauthor{\bsnm{Bang}, \binits{J.}},
\bauthor{\bsnm{Kim}, \binits{H.}},
\bauthor{\bsnm{Yoo}, \binits{Y.}},
\bauthor{\bsnm{Ha}, \binits{J.-W.}},
\bauthor{\bsnm{Choi}, \binits{J.}}:
\bctitle{Rainbow memory: Continual learning with a memory of diverse samples}.
In: \bbtitle{Proceedings of the IEEE/CVF Conference on Computer Vision and
  Pattern Recognition},
pp. \bfpage{8218}--\blpage{8227}
(\byear{2021})
\end{bchapter}
\endbibitem

\bibitem{eysenbach2018diversity}
\begin{bchapter}
\bauthor{\bsnm{Eysenbach}, \binits{B.}},
\bauthor{\bsnm{Gupta}, \binits{A.}},
\bauthor{\bsnm{Ibarz}, \binits{J.}},
\bauthor{\bsnm{Levine}, \binits{S.}}:
\bctitle{Diversity is all you need: Learning skills without a reward function}.
In: \bbtitle{International Conference on Learning Representations}
(\byear{2018})
\end{bchapter}
\endbibitem

\bibitem{parker2020effective}
\begin{botherref}
\oauthor{\bsnm{Parker-Holder}, \binits{J.}},
\oauthor{\bsnm{Pacchiano}, \binits{A.}},
\oauthor{\bsnm{Choromanski}, \binits{K.M.}},
\oauthor{\bsnm{Roberts}, \binits{S.J.}}:
Effective diversity in population based reinforcement learning.
Advances in Neural Information Processing Systems
\textbf{33}
(2020)
\end{botherref}
\endbibitem

\bibitem{dai2021diversity}
\begin{bchapter}
\bauthor{\bsnm{Dai}, \binits{T.}},
\bauthor{\bsnm{Liu}, \binits{H.}},
\bauthor{\bsnm{Arulkumaran}, \binits{K.}},
\bauthor{\bsnm{Ren}, \binits{G.}},
\bauthor{\bsnm{Bharath}, \binits{A.A.}}:
\bctitle{Diversity-based trajectory and goal selection with hindsight
  experience replay}.
In: \bbtitle{Pacific Rim International Conference on Artificial Intelligence},
pp. \bfpage{32}--\blpage{45}
(\byear{2021}).
\bcomment{Springer}
\end{bchapter}
\endbibitem

\bibitem{Breiman1996}
\begin{barticle}
\bauthor{\bsnm{Breiman}, \binits{L.}}:
\batitle{{B}agging {P}redictors}.
\bjtitle{{{Machine Learning}}}
\bvolume{24},
\bfpage{123}--\blpage{140}
(\byear{1996}).
\doiurl{10.1023/A:1018054314350}
\end{barticle}
\endbibitem

\bibitem{glorot2010understanding}
\begin{bchapter}
\bauthor{\bsnm{Glorot}, \binits{X.}},
\bauthor{\bsnm{Bengio}, \binits{Y.}}:
\bctitle{Understanding the difficulty of training deep feedforward neural
  networks}.
In: \beditor{\bsnm{Teh}, \binits{Y.W.}},
\beditor{\bsnm{Titterington}, \binits{M.}} (eds.)
\bbtitle{Proceedings of the Thirteenth International Conference on Artificial
  Intelligence and Statistics}.
\bsertitle{Proceedings of Machine Learning Research},
vol. \bseriesno{9},
pp. \bfpage{249}--\blpage{256}.
\bpublisher{PMLR},
\blocation{Chia Laguna Resort, Sardinia, Italy}
(\byear{2010}).
\burl{https://proceedings.mlr.press/v9/glorot10a.html}
\end{bchapter}
\endbibitem

\bibitem{he2015delving}
\begin{bchapter}
\bauthor{\bsnm{He}, \binits{K.}},
\bauthor{\bsnm{Zhang}, \binits{X.}},
\bauthor{\bsnm{Ren}, \binits{S.}},
\bauthor{\bsnm{Sun}, \binits{J.}}:
\bctitle{Delving deep into rectifiers: Surpassing human-level performance on
  imagenet classification}.
In: \bbtitle{Proceedings of the IEEE International Conference on Computer
  Vision},
pp. \bfpage{1026}--\blpage{1034}
(\byear{2015})
\end{bchapter}
\endbibitem

\bibitem{narkhede2021review}
\begin{botherref}
\oauthor{\bsnm{Narkhede}, \binits{M.V.}},
\oauthor{\bsnm{Bartakke}, \binits{P.P.}},
\oauthor{\bsnm{Sutaone}, \binits{M.S.}}:
A review on weight initialization strategies for neural networks.
Artificial intelligence review,
1--32
(2021)
\end{botherref}
\endbibitem

\bibitem{Genewein2015}
\begin{barticle}
\bauthor{\bsnm{Genewein}, \binits{T.}},
\bauthor{\bsnm{Leibfried}, \binits{F.}},
\bauthor{\bsnm{Grau-Moya}, \binits{J.}},
\bauthor{\bsnm{Braun}, \binits{D.A.}}:
\batitle{Bounded rationality, abstraction, and hierarchical decision-making: An
  information-theoretic optimality principle}.
\bjtitle{Frontiers in Robotics and AI}
\bvolume{2},
\bfpage{27}
(\byear{2015})
\end{barticle}
\endbibitem

\bibitem{galashov2019information}
\begin{bchapter}
\bauthor{\bsnm{Galashov}, \binits{A.}},
\bauthor{\bsnm{Jayakumar}, \binits{S.M.}},
\bauthor{\bsnm{Hasenclever}, \binits{L.}},
\bauthor{\bsnm{Tirumala}, \binits{D.}},
\bauthor{\bsnm{Schwarz}, \binits{J.}},
\bauthor{\bsnm{Desjardins}, \binits{G.}},
\bauthor{\bsnm{Czarnecki}, \binits{W.M.}},
\bauthor{\bsnm{Teh}, \binits{Y.W.}},
\bauthor{\bsnm{Pascanu}, \binits{R.}},
\bauthor{\bsnm{Heess}, \binits{N.}}:
\bctitle{Information asymmetry in kl-regularized rl}.
In: \bbtitle{Proceedings of the International Conference on Representation
  Learning}
(\byear{2019})
\end{bchapter}
\endbibitem

\bibitem{grau2018soft}
\begin{bchapter}
\bauthor{\bsnm{Grau-Moya}, \binits{J.}},
\bauthor{\bsnm{Leibfried}, \binits{F.}},
\bauthor{\bsnm{Vrancx}, \binits{P.}}:
\bctitle{Soft q-learning with mutual-information regularization}.
In: \bbtitle{Proceedings of the International Conference on Learning
  Representations}
(\byear{2019})
\end{bchapter}
\endbibitem

\bibitem{macchi1975coincidence}
\begin{barticle}
\bauthor{\bsnm{Macchi}, \binits{O.}}:
\batitle{The coincidence approach to stochastic point processes}.
\bjtitle{Advances in Applied Probability}
\bvolume{7}(\bissue{1}),
\bfpage{83}--\blpage{122}
(\byear{1975})
\end{barticle}
\endbibitem

\bibitem{cover2012elements}
\begin{bbook}
\bauthor{\bsnm{Cover}, \binits{T.M.}},
\bauthor{\bsnm{Thomas}, \binits{J.A.}}:
\bbtitle{Elements of Information Theory}.
\bpublisher{John Wiley \& Sons},
\blocation{Hoboken, NJ}
(\byear{2012})
\end{bbook}
\endbibitem

\bibitem{ebrahimi2020uncertainty}
\begin{bchapter}
\bauthor{\bsnm{Ebrahimi}, \binits{S.}},
\bauthor{\bsnm{Elhoseiny}, \binits{M.}},
\bauthor{\bsnm{Darrell}, \binits{T.}},
\bauthor{\bsnm{Rohrbach}, \binits{M.}}:
\bctitle{Uncertainty-guided continual learning with bayesian neural networks}.
In: \bbtitle{International Conference on Learning Representations}
(\byear{2020})
\end{bchapter}
\endbibitem

\bibitem{van2020brain}
\begin{barticle}
\bauthor{\bparticle{van~de} \bsnm{Ven}, \binits{G.M.}},
\bauthor{\bsnm{Siegelmann}, \binits{H.T.}},
\bauthor{\bsnm{Tolias}, \binits{A.S.}}:
\batitle{Brain-inspired replay for continual learning with artificial neural
  networks}.
\bjtitle{Nature communications}
\bvolume{11}(\bissue{1}),
\bfpage{1}--\blpage{14}
(\byear{2020})
\end{barticle}
\endbibitem

\bibitem{kessler2021hierarchical}
\begin{bchapter}
\bauthor{\bsnm{Kessler}, \binits{S.}},
\bauthor{\bsnm{Nguyen}, \binits{V.}},
\bauthor{\bsnm{Zohren}, \binits{S.}},
\bauthor{\bsnm{Roberts}, \binits{S.J.}}:
\bctitle{Hierarchical indian buffet neural networks for bayesian continual
  learning}.
In: \bbtitle{Uncertainty in Artificial Intelligence},
pp. \bfpage{749}--\blpage{759}
(\byear{2021}).
\bcomment{PMLR}
\end{bchapter}
\endbibitem

\bibitem{raghavan2021formalizing}
\begin{botherref}
\oauthor{\bsnm{Raghavan}, \binits{K.}},
\oauthor{\bsnm{Balaprakash}, \binits{P.}}:
Formalizing the generalization-forgetting trade-off in continual learning.
Advances in Neural Information Processing Systems
\textbf{34}
(2021)
\end{botherref}
\endbibitem

\bibitem{mazur2021target}
\begin{botherref}
\oauthor{\bsnm{Mazur}, \binits{M.}},
\oauthor{\bsnm{Pustelnik}, \binits{{\L}.}},
\oauthor{\bsnm{Knop}, \binits{S.}},
\oauthor{\bsnm{Pagacz}, \binits{P.}},
\oauthor{\bsnm{Spurek}, \binits{P.}}:
Target layer regularization for continual learning using cramer-wold generator.
arXiv preprint arXiv:2111.07928
(2021)
\end{botherref}
\endbibitem

\bibitem{van2018generative}
\begin{bchapter}
\bauthor{\bparticle{van~de} \bsnm{Ven}, \binits{G.M.}},
\bauthor{\bsnm{Tolias}, \binits{A.S.}}:
\bctitle{Generative replay with feedback connections as a general strategy for
  continual learning}.
In: \bbtitle{arXiv Preprint arXiv:1809.10635}
(\byear{2018})
\end{bchapter}
\endbibitem

\bibitem{farquhar2018towards}
\begin{bchapter}
\bauthor{\bsnm{Farquhar}, \binits{S.}},
\bauthor{\bsnm{Gal}, \binits{Y.}}:
\bctitle{Towards robust evaluations of continual learning}.
In: \bbtitle{Lifelong Learning: A Reinforcement Learning Approach (ICML 2018)}
(\byear{2018})
\end{bchapter}
\endbibitem

\bibitem{hsu2018re}
\begin{bchapter}
\bauthor{\bsnm{Hsu}, \binits{Y.-C.}},
\bauthor{\bsnm{Liu}, \binits{Y.-C.}},
\bauthor{\bsnm{Ramasamy}, \binits{A.}},
\bauthor{\bsnm{Kira}, \binits{Z.}}:
\bctitle{Re-evaluating continual learning scenarios: A categorization and case
  for strong baselines}.
In: \bbtitle{Continual Learning Workshop, 32nd Conference on Neural Information
  Processing Systems}
(\byear{2018})
\end{bchapter}
\endbibitem

\bibitem{he2022online}
\begin{bchapter}
\bauthor{\bsnm{He}, \binits{J.}},
\bauthor{\bsnm{Zhu}, \binits{F.}}:
\bctitle{Online continual learning via candidates voting}.
In: \bbtitle{Proceedings of the IEEE/CVF Winter Conference on Applications of
  Computer Vision},
pp. \bfpage{3154}--\blpage{3163}
(\byear{2022})
\end{bchapter}
\endbibitem

\bibitem{kao2021natural}
\begin{botherref}
\oauthor{\bsnm{Kao}, \binits{T.-C.}},
\oauthor{\bsnm{Jensen}, \binits{K.}},
\oauthor{\bparticle{van~de} \bsnm{Ven}, \binits{G.}},
\oauthor{\bsnm{Bernacchia}, \binits{A.}},
\oauthor{\bsnm{Hennequin}, \binits{G.}}:
Natural continual learning: success is a journey, not (just) a destination.
Advances in Neural Information Processing Systems
\textbf{34}
(2021)
\end{botherref}
\endbibitem

\bibitem{hadjeres2017glsr}
\begin{bchapter}
\bauthor{\bsnm{Hadjeres}, \binits{G.}},
\bauthor{\bsnm{Nielsen}, \binits{F.}},
\bauthor{\bsnm{Pachet}, \binits{F.}}:
\bctitle{Glsr-vae: Geodesic latent space regularization for variational
  autoencoder architectures}.
In: \bbtitle{2017 IEEE Symposium Series on Computational Intelligence (SSCI)},
pp. \bfpage{1}--\blpage{7}
(\byear{2017}).
\bcomment{IEEE}
\end{bchapter}
\endbibitem

\bibitem{ghosh2019variational}
\begin{bchapter}
\bauthor{\bsnm{Ghosh}, \binits{P.}},
\bauthor{\bsnm{Sajjadi}, \binits{M.S.}},
\bauthor{\bsnm{Vergari}, \binits{A.}},
\bauthor{\bsnm{Black}, \binits{M.}},
\bauthor{\bsnm{Scholkopf}, \binits{B.}}:
\bctitle{From variational to deterministic autoencoders}.
In: \bbtitle{International Conference on Learning Representations}
(\byear{2019})
\end{bchapter}
\endbibitem

\bibitem{vahdat2020nvae}
\begin{bchapter}
\bauthor{\bsnm{Vahdat}, \binits{A.}},
\bauthor{\bsnm{Kautz}, \binits{J.}}:
\bctitle{Nvae: A deep hierarchical variational autoencoder}.
In: \beditor{\bsnm{Larochelle}, \binits{H.}},
\beditor{\bsnm{Ranzato}, \binits{M.}},
\beditor{\bsnm{Hadsell}, \binits{R.}},
\beditor{\bsnm{Balcan}, \binits{M.F.}},
\beditor{\bsnm{Lin}, \binits{H.}} (eds.)
\bbtitle{Advances in Neural Information Processing Systems},
vol. \bseriesno{33},
pp. \bfpage{19667}--\blpage{19679}.
\bpublisher{Curran Associates, Inc.},
\blocation{Online Conference}
(\byear{2020}).
\burl{https://proceedings.neurips.cc/paper/2020/file/e3b21256183cf7c2c7a66be163579d37-Paper.pdf}
\end{bchapter}
\endbibitem

\bibitem{coumans2021pybullet}
\begin{botherref}
\oauthor{\bsnm{Coumans}, \binits{E.}},
\oauthor{\bsnm{Bai}, \binits{Y.}}:
PyBullet, a Python module for physics simulation for games, robotics and
  machine learning.
\url{http://pybullet.org}
(2016--2021)
\end{botherref}
\endbibitem

\bibitem{benelot2018pybulletgym}
\begin{botherref}
\oauthor{\bsnm{Ellenberger}, \binits{B.}}:
PyBullet Gymperium.
\url{ https://github.com/benelot/pybullet-gym}
(2018--2019)
\end{botherref}
\endbibitem

\bibitem{wang2020deep}
\begin{botherref}
\oauthor{\bsnm{Wang}, \binits{H.-n.}},
\oauthor{\bsnm{Liu}, \binits{N.}},
\oauthor{\bsnm{Zhang}, \binits{Y.-y.}},
\oauthor{\bsnm{Feng}, \binits{D.-w.}},
\oauthor{\bsnm{Huang}, \binits{F.}},
\oauthor{\bsnm{Li}, \binits{D.-s.}},
\oauthor{\bsnm{Zhang}, \binits{Y.-m.}}:
Deep reinforcement learning: a survey.
Frontiers of Information Technology \& Electronic Engineering,
1--19
(2020)
\end{botherref}
\endbibitem

\bibitem{haarnoja2018soft}
\begin{bchapter}
\bauthor{\bsnm{Haarnoja}, \binits{T.}},
\bauthor{\bsnm{Zhou}, \binits{A.}},
\bauthor{\bsnm{Abbeel}, \binits{P.}},
\bauthor{\bsnm{Levine}, \binits{S.}}:
\bctitle{Soft actor-critic: Off-policy maximum entropy deep reinforcement
  learning with a stochastic actor}.
In: \bbtitle{International Conference on Machine Learning},
pp. \bfpage{1861}--\blpage{1870}
(\byear{2018})
\end{bchapter}
\endbibitem

\bibitem{salimans2016improved}
\begin{barticle}
\bauthor{\bsnm{Salimans}, \binits{T.}},
\bauthor{\bsnm{Goodfellow}, \binits{I.}},
\bauthor{\bsnm{Zaremba}, \binits{W.}},
\bauthor{\bsnm{Cheung}, \binits{V.}},
\bauthor{\bsnm{Radford}, \binits{A.}},
\bauthor{\bsnm{Chen}, \binits{X.}}:
\batitle{Improved techniques for training gans}.
\bjtitle{Advances in neural information processing systems}
\bvolume{29},
\bfpage{2234}--\blpage{2242}
(\byear{2016})
\end{barticle}
\endbibitem

\bibitem{szegedy2015going}
\begin{bchapter}
\bauthor{\bsnm{Szegedy}, \binits{C.}},
\bauthor{\bsnm{Liu}, \binits{W.}},
\bauthor{\bsnm{Jia}, \binits{Y.}},
\bauthor{\bsnm{Sermanet}, \binits{P.}},
\bauthor{\bsnm{Reed}, \binits{S.}},
\bauthor{\bsnm{Anguelov}, \binits{D.}},
\bauthor{\bsnm{Erhan}, \binits{D.}},
\bauthor{\bsnm{Vanhoucke}, \binits{V.}},
\bauthor{\bsnm{Rabinovich}, \binits{A.}}:
\bctitle{Going deeper with convolutions}.
In: \bbtitle{Proceedings of the IEEE Conference on Computer Vision and Pattern
  Recognition},
pp. \bfpage{1}--\blpage{9}
(\byear{2015})
\end{bchapter}
\endbibitem

\bibitem{fu2019cyclical}
\begin{bchapter}
\bauthor{\bsnm{Fu}, \binits{H.}},
\bauthor{\bsnm{Li}, \binits{C.}},
\bauthor{\bsnm{Liu}, \binits{X.}},
\bauthor{\bsnm{Gao}, \binits{J.}},
\bauthor{\bsnm{Celikyilmaz}, \binits{A.}},
\bauthor{\bsnm{Carin}, \binits{L.}}:
\bctitle{Cyclical annealing schedule: A simple approach to mitigating kl
  vanishing}.
In: \bbtitle{Proceedings of the 2019 Conference of the North American Chapter
  of the Association for Computational Linguistics: Human Language
  Technologies, Volume 1 (Long and Short Papers)},
pp. \bfpage{240}--\blpage{250}
(\byear{2019})
\end{bchapter}
\endbibitem

\bibitem{ghosh2018divide}
\begin{bchapter}
\bauthor{\bsnm{Ghosh}, \binits{D.}},
\bauthor{\bsnm{Singh}, \binits{A.}},
\bauthor{\bsnm{Rajeswaran}, \binits{A.}},
\bauthor{\bsnm{Kumar}, \binits{V.}},
\bauthor{\bsnm{Levine}, \binits{S.}}:
\bctitle{Divide-and-conquer reinforcement learning}.
In: \bbtitle{International Conference on Learning Representations}
(\byear{2018})
\end{bchapter}
\endbibitem

\bibitem{leibfried2019mutual}
\begin{bchapter}
\bauthor{\bsnm{Leibfried}, \binits{F.}},
\bauthor{\bsnm{Grau-Moya}, \binits{J.}}:
\bctitle{Mutual-information regularization in markov decision processes and
  actor-critic learning}.
In: \bbtitle{Proceedings of the Conference on Robot Learning}
(\byear{2019})
\end{bchapter}
\endbibitem

\bibitem{hihn2019information}
\begin{bchapter}
\bauthor{\bsnm{Hihn}, \binits{H.}},
\bauthor{\bsnm{Gottwald}, \binits{S.}},
\bauthor{\bsnm{Braun}, \binits{D.A.}}:
\bctitle{An information-theoretic on-line learning principle for specialization
  in hierarchical decision-making systems}.
In: \bbtitle{2019 IEEE 58th Conference on Decision and Control (CDC)},
pp. \bfpage{3677}--\blpage{3684}
(\byear{2019}).
\bcomment{IEEE}
\end{bchapter}
\endbibitem

\bibitem{arumugam2021information}
\begin{bchapter}
\bauthor{\bsnm{Arumugam}, \binits{D.}},
\bauthor{\bsnm{Henderson}, \binits{P.}},
\bauthor{\bsnm{Bacon}, \binits{P.-L.}}:
\bctitle{An information-theoretic perspective on credit assignment in
  reinforcement learning}.
In: \bbtitle{Workshop on Biological and Artificial Reinforcement Learning
  (NeurIPS 2020)}
(\byear{2020})
\end{bchapter}
\endbibitem

\bibitem{hihn2018bounded}
\begin{bchapter}
\bauthor{\bsnm{Hihn}, \binits{H.}},
\bauthor{\bsnm{Gottwald}, \binits{S.}},
\bauthor{\bsnm{Braun}, \binits{D.A.}}:
\bctitle{Bounded rational decision-making with adaptive neural network priors}.
In: \bbtitle{IAPR Workshop on Artificial Neural Networks in Pattern
  Recognition},
pp. \bfpage{213}--\blpage{225}
(\byear{2018}).
\bcomment{Springer}
\end{bchapter}
\endbibitem

\bibitem{pang2020learning}
\begin{botherref}
\oauthor{\bsnm{Pang}, \binits{B.}},
\oauthor{\bsnm{Han}, \binits{T.}},
\oauthor{\bsnm{Nijkamp}, \binits{E.}},
\oauthor{\bsnm{Zhu}, \binits{S.-C.}},
\oauthor{\bsnm{Wu}, \binits{Y.N.}}:
Learning latent space energy-based prior model.
Advances in Neural Information Processing Systems
\textbf{33}
(2020)
\end{botherref}
\endbibitem

\bibitem{rothfuss2018promp}
\begin{bchapter}
\bauthor{\bsnm{Rothfuss}, \binits{J.}},
\bauthor{\bsnm{Lee}, \binits{D.}},
\bauthor{\bsnm{Clavera}, \binits{I.}},
\bauthor{\bsnm{Asfour}, \binits{T.}},
\bauthor{\bsnm{Abbeel}, \binits{P.}}:
\bctitle{Promp: Proximal meta-policy search}.
In: \bbtitle{International Conference on Learning Representations}
(\byear{2018})
\end{bchapter}
\endbibitem

\bibitem{hihn2020hierarchical}
\begin{bchapter}
\bauthor{\bsnm{Hihn}, \binits{H.}},
\bauthor{\bsnm{Braun}, \binits{D.A.}}:
\bctitle{Hierarchical expert networks for meta-learning}.
In: \bbtitle{4th ICML Workshop on Life Long Machine Learning}
(\byear{2020})
\end{bchapter}
\endbibitem

\bibitem{thiam2021multi}
\begin{botherref}
\oauthor{\bsnm{Thiam}, \binits{P.}},
\oauthor{\bsnm{Hihn}, \binits{H.}},
\oauthor{\bsnm{Braun}, \binits{D.A.}},
\oauthor{\bsnm{Kestler}, \binits{H.A.}},
\oauthor{\bsnm{Schwenker}, \binits{F.}}:
Multi-modal pain intensity assessment based on physiological signals: A deep
  learning perspective.
Frontiers in Physiology
\textbf{12}
(2021)
\end{botherref}
\endbibitem

\bibitem{tsai2021self}
\begin{bchapter}
\bauthor{\bsnm{Tsai}, \binits{Y.-H.H.}},
\bauthor{\bsnm{Wu}, \binits{Y.}},
\bauthor{\bsnm{Salakhutdinov}, \binits{R.}},
\bauthor{\bsnm{Morency}, \binits{L.-P.}}:
\bctitle{Self-supervised learning from a multi-view perspective}.
In: \bbtitle{International Conference on Learning Representations}
(\byear{2021})
\end{bchapter}
\endbibitem

\bibitem{antoniak1974mixtures}
\begin{botherref}
\oauthor{\bsnm{Antoniak}, \binits{C.E.}}:
Mixtures of dirichlet processes with applications to bayesian nonparametric
  problems.
The annals of statistics,
1152--1174
(1974)
\end{botherref}
\endbibitem

\bibitem{lupu2021trajectory}
\begin{bchapter}
\bauthor{\bsnm{Lupu}, \binits{A.}},
\bauthor{\bsnm{Cui}, \binits{B.}},
\bauthor{\bsnm{Hu}, \binits{H.}},
\bauthor{\bsnm{Foerster}, \binits{J.}}:
\bctitle{Trajectory diversity for zero-shot coordination}.
In: \bbtitle{International Conference on Machine Learning},
pp. \bfpage{7204}--\blpage{7213}
(\byear{2021}).
\bcomment{PMLR}
\end{bchapter}
\endbibitem

\bibitem{andrychowicz2017hindsight}
\begin{bchapter}
\bauthor{\bsnm{Andrychowicz}, \binits{M.}},
\bauthor{\bsnm{Wolski}, \binits{F.}},
\bauthor{\bsnm{Ray}, \binits{A.}},
\bauthor{\bsnm{Schneider}, \binits{J.}},
\bauthor{\bsnm{Fong}, \binits{R.}},
\bauthor{\bsnm{Welinder}, \binits{P.}},
\bauthor{\bsnm{McGrew}, \binits{B.}},
\bauthor{\bsnm{Tobin}, \binits{J.}},
\bauthor{\bsnm{Abbeel}, \binits{P.}},
\bauthor{\bsnm{Zaremba}, \binits{W.}}:
\bctitle{Hindsight experience replay}.
In: \bbtitle{Proceedings of the 31st International Conference on Neural
  Information Processing Systems},
pp. \bfpage{5055}--\blpage{5065}
(\byear{2017})
\end{bchapter}
\endbibitem

\bibitem{jung2020continual}
\begin{barticle}
\bauthor{\bsnm{Jung}, \binits{S.}},
\bauthor{\bsnm{Ahn}, \binits{H.}},
\bauthor{\bsnm{Cha}, \binits{S.}},
\bauthor{\bsnm{Moon}, \binits{T.}}:
\batitle{Continual learning with node-importance based adaptive group sparse
  regularization}.
\bjtitle{Advances in Neural Information Processing Systems}
\bvolume{33},
\bfpage{3647}--\blpage{3658}
(\byear{2020})
\end{barticle}
\endbibitem

\bibitem{gal2016bayesian}
\begin{bchapter}
\bauthor{\bsnm{Gal}, \binits{Y.}},
\bauthor{\bsnm{Ghahramani}, \binits{Z.}}:
\bctitle{Bayesian convolutional neural networks with bernoulli approximate
  variational inference}.
In: \bbtitle{ICLR 2016 Workshop Track}
(\byear{2016})
\end{bchapter}
\endbibitem

\bibitem{zhang2018noisy}
\begin{bchapter}
\bauthor{\bsnm{Zhang}, \binits{G.}},
\bauthor{\bsnm{Sun}, \binits{S.}},
\bauthor{\bsnm{Duvenaud}, \binits{D.}},
\bauthor{\bsnm{Grosse}, \binits{R.}}:
\bctitle{Noisy natural gradient as variational inference}.
In: \bbtitle{International Conference on Machine Learning},
pp. \bfpage{5852}--\blpage{5861}
(\byear{2018}).
\bcomment{PMLR}
\end{bchapter}
\endbibitem

\bibitem{Freitas2000sequential}
\begin{barticle}
\bauthor{\bsnm{Freitas}, \binits{J.d.}},
\bauthor{\bsnm{Niranjan}, \binits{M.}},
\bauthor{\bsnm{Gee}, \binits{A.H.}},
\bauthor{\bsnm{Doucet}, \binits{A.}}:
\batitle{Sequential monte carlo methods to train neural network models}.
\bjtitle{Neural computation}
\bvolume{12}(\bissue{4}),
\bfpage{955}--\blpage{993}
(\byear{2000})
\end{barticle}
\endbibitem

\bibitem{wen2018flipout}
\begin{bchapter}
\bauthor{\bsnm{Wen}, \binits{Y.}},
\bauthor{\bsnm{Vicol}, \binits{P.}},
\bauthor{\bsnm{Ba}, \binits{J.}},
\bauthor{\bsnm{Tran}, \binits{D.}},
\bauthor{\bsnm{Grosse}, \binits{R.}}:
\bctitle{Flipout: Efficient pseudo-independent weight perturbations on
  mini-batches}.
In: \bbtitle{International Conference on Learning Representations}
(\byear{2018})
\end{bchapter}
\endbibitem

\bibitem{lazyadam}
\begin{botherref}
Tensorflow 2.0 Documentation
(2022).
\url{https://www.tensorflow.org/addons/api_docs/python/tfa/optimizers/LazyAdam}
\end{botherref}
\endbibitem

\bibitem{jerfel2019reconciling}
\begin{bchapter}
\bauthor{\bsnm{Jerfel}, \binits{G.}},
\bauthor{\bsnm{Grant}, \binits{E.}},
\bauthor{\bsnm{Griffiths}, \binits{T.}},
\bauthor{\bsnm{Heller}, \binits{K.A.}}:
\bctitle{Reconciling meta-learning and continual learning with online mixtures
  of tasks}.
In: \bbtitle{Advances in Neural Information Processing Systems},
pp. \bfpage{9122}--\blpage{9133}
(\byear{2019})
\end{bchapter}
\endbibitem

\bibitem{maas2013rectifier}
\begin{bchapter}
\bauthor{\bsnm{Maas}, \binits{A.L.}},
\bauthor{\bsnm{Hannun}, \binits{A.Y.}},
\bauthor{\bsnm{Ng}, \binits{A.Y.}}:
\bctitle{Rectifier nonlinearities improve neural network acoustic models}.
In: \bbtitle{Proc. Icml},
vol. \bseriesno{30},
p. \bfpage{3}
(\byear{2013})
\end{bchapter}
\endbibitem

\bibitem{srivastava2014dropout}
\begin{barticle}
\bauthor{\bsnm{Srivastava}, \binits{N.}},
\bauthor{\bsnm{Hinton}, \binits{G.}},
\bauthor{\bsnm{Krizhevsky}, \binits{A.}},
\bauthor{\bsnm{Sutskever}, \binits{I.}},
\bauthor{\bsnm{Salakhutdinov}, \binits{R.}}:
\batitle{Dropout: a simple way to prevent neural networks from overfitting}.
\bjtitle{The journal of machine learning research}
\bvolume{15}(\bissue{1}),
\bfpage{1929}--\blpage{1958}
(\byear{2014})
\end{barticle}
\endbibitem

\bibitem{KingmaBa2015}
\begin{bchapter}
\bauthor{\bsnm{Kingma}, \binits{D.P.}},
\bauthor{\bsnm{Ba}, \binits{J.}}:
\bctitle{{A}dam: {A} method for stochastic optimization}.
In: \bbtitle{Proceedings of the 3rd International Conference on Learning
  Representations}
(\byear{2015})
\end{bchapter}
\endbibitem

\bibitem{lopez2017gradient}
\begin{bchapter}
\bauthor{\bsnm{Lopez-Paz}, \binits{D.}},
\bauthor{\bsnm{Ranzato}, \binits{M.}}:
\bctitle{Gradient episodic memory for continual learning}.
In: \bbtitle{Advances in Neural Information Processing Systems},
pp. \bfpage{6467}--\blpage{6476}
(\byear{2017})
\end{bchapter}
\endbibitem

\bibitem{kj2020meta}
\begin{barticle}
\bauthor{\bsnm{KJ}, \binits{J.}},
\bauthor{\bsnm{N~Balasubramanian}, \binits{V.}}:
\batitle{Meta-consolidation for continual learning}.
\bjtitle{Advances in Neural Information Processing Systems}
\bvolume{33},
\bfpage{14374}--\blpage{14386}
(\byear{2020})
\end{barticle}
\endbibitem

\bibitem{schaul2015prioritized}
\begin{botherref}
\oauthor{\bsnm{Schaul}, \binits{T.}},
\oauthor{\bsnm{Quan}, \binits{J.}},
\oauthor{\bsnm{Antonoglou}, \binits{I.}},
\oauthor{\bsnm{Silver}, \binits{D.}}:
Prioritized experience replay.
arXiv preprint arXiv:1511.05952
(2015)
\end{botherref}
\endbibitem

\bibitem{Schulman2017}
\begin{bchapter}
\bauthor{\bsnm{Schulman}, \binits{J.}},
\bauthor{\bsnm{Wolski}, \binits{F.}},
\bauthor{\bsnm{Dhariwal}, \binits{P.}},
\bauthor{\bsnm{Radford}, \binits{A.}},
\bauthor{\bsnm{Klimov}, \binits{O.}}:
\bctitle{Proximal policy optimization algorithms}.
In: \bbtitle{arXiv Preprint arXiv:1707.06347}
(\byear{2017})
\end{bchapter}
\endbibitem

\end{thebibliography}


\begin{appendices}

\section{Wasserstein-Distance between two Gaussians}
\label{app:wassergaussian}
The $W_2^2$ distance between two Gaussians is given by 
\begin{equation}
W_2^2(p,q) = \lVert \mu_p - \mu_q\rVert_2^2 + \lVert\sqrt{d_p} - \sqrt{d_q}\rVert_2,
\end{equation}

\textit{Proof:} Let $p = \mathcal{N}(\mu_p, \Sigma_p)$ and $q = \mathcal{N}(\mu_q, \sigma_q)$ be two Guassian distributions. The Wasserstein-2 distance between $p$ and $q$ is then given by
\begin{equation}
W_2^2(p,q) = \lVert\mu_p - \mu_q\rVert_2^2 + \mathcal{B}(\Sigma_p,\Sigma_q),
\end{equation}
where $\mathcal{B}$ is the Bures metric between two positive semi-definite matrices:
\begin{equation}
\mathcal{B}(\Sigma_p,\Sigma_q) = tr(\Sigma_p + \Sigma_q - 2(\Sigma_p^{1/2} \Sigma_q \Sigma_p^{1/2})^{1/2},
\end{equation}
where $tr(A)$ is the trace of a matrix $A$ and $A^{1/2}$ is the matrix square root. Matrix square roots are computationally expensive to compute and there can potentially be an infinite number of solutions. In the case where $p$ and $q$ are Gaussian mean-field approximations, i.e., all dimensions are independent, $\Sigma_p$ and $\Sigma_q$ are given by diagonal matrices, such that $\Sigma_p = \text{diag}(d_p)_i$ and $\Sigma_q = \text{diag}(d_q)_i$. The Bures metric then reduces to the Hellinger distance between the diagonals $d_p$ and $d_q$, and we have:
\begin{equation}
W_2^2(p,q) = \lVert\mu_p - \mu_q\rVert_2^2 + \lVert\sqrt{d_p} - \sqrt{d_q}\rVert_2.
\label{eq:wassernorm}
\end{equation}
The full Hellinger distance is given by  $\frac{1}{\sqrt{2}}\lVert\sqrt{d_p} - \sqrt{d_q}\rVert_2$, but we chose to ommit the constant  factor during optimization.
\section{Wasserstein-2 Exponential Kernel}
\label{app:wasserkernel}
The exponential Wasserstein-2 kernel between isotropic Gaussian distributions $p$ and $q$ with kernel width $h$ defined by
\begin{equation*}
k(p,q) = \exp\left(-\frac{W_2^2(p,q)}{2h^2}\right)
\end{equation*} is a valid kernel function.

\textit{Proof:} The simplest way to show a kernel function $k$ is valid is by deriving $k$ from other valid kernels. We can express the Wasserstein distance as the sum of two norms as shown in Equation \eqref{eq:wassernorm}. The euclidean norm and the Hellinger distance both form inner product spaces and are thus valid kernel functions. Their sum is also a valid kernel function, which makes the Wasserstein distance on isotropic Gaussians a valid kernel. If $k(p,q)$ is a valid kernel, then $\exp(k(p,q))$ is also a valid kernel.

\section{Experiment Details}
\label{app:cifarexperiments}

To implement variational layers we use Gaussian distributions. For simplicity we use a $D-$dimensional Gaussian mean-field approximate posterior $q_t(\theta)= \prod^D_{d=1}\mathcal{N}(\theta_t\vert p_{t,d},\sigma^2_{t,d})$. We use the flip-out estimator \cite{wen2018flipout} to approximate the gradients. In practice, we draw a single sample to approximate the expectation. 
\subsection{MNIST Experiments}
For split MNIST experiments we used dense layers for both the VAE and the classifier. The VAE encoder contains two layers with 256 units each, followed by 64 units (64 units for the mean and 64 units for log-variance) for the latent variable, and two layers with 256 units for the decoder, followed by an output layer with $28*28 = 784$ units. This assumes isotropic Gaussians as priors and posteriors over the latent variable and allows to compute the $\DKL$ if closed form. We used only one expert for the VAE with $\beta_1 = 0.002$, $\beta_2 = 0.75$, a diversity bonus weight of $0.01$ and  leaky ReLU activations \cite{maas2013rectifier} in the hidden layers. We trained with a batch size 256 for 150 epochs. The VAE output activation function is a sigmoid and we trained it using a binary cross-entropy loss between the normalized pixel values of the original and the reconstructed images.  We used no other regularization methods on the VAE. We used 10.000 generated samples after each task.

The classifier consists of two dense layers, each with 256 units with leaky ReLU activations \cite{maas2013rectifier} and dropout \cite{srivastava2014dropout} layers, followed by an output layer with two units. All layers of the classifier have two experts. We trained with batch size 256 for 150 epochs using Adam \cite{KingmaBa2015} with a learning rate of $6*10^{-4}$. In the permuted MNIST setting we used the same architecture, but increased the number of units to 512.
\begin{table}[t!]
\small
\setlength{\tabcolsep}{2pt} 
\centering
\begin{tabularx}{\columnwidth}{*{3}l}
\toprule
\textbf{Baselines} &  \textbf{S-MNIST} & \textbf{P-MNIST} \\
\midrule
Dense Neural Network &  86.15 ($\pm$1.00)  & 17.26 ($\pm$0.19) \\
Offline re-training + task oracle &  99.64 ($\pm$0.03)& 97.59 ($\pm$0.02) \\
\midrule
\textbf{Single-Head and Task-Agnostic Methods} & & \\
\midrule
Hierarchical VCL (ours) & 97.50 ($\pm0.33$) & 97.07 ($\pm$0.62) \\
Hierarchical VCL w/ GR (ours) & 98.60 ($\pm0.35$) & 97.47 ($\pm$0.52) \\
Uncertainty Guided CL w/ BNN \cite{ebrahimi2020uncertainty} & 97.70 ($\pm$0.03) & 92.50 ($\pm$0.01) \\
Brain-inspired Replay through Feedback$^\dagger$ \cite{van2020brain} & 99.66 ($\pm$0.13) & 97.31 ($\pm$0.04) \\
Hierarchical Indian Buffet Neural Nets \cite{kessler2021hierarchical} & 91.00 ($\pm$2.20) & 93.70 ($\pm$0.60)\\
Balanced Continual Learning \cite{raghavan2021formalizing} & 98.71 ($\pm$0.06) & 97.51 ($\pm$0.05) \\
Target Layer Regularization \cite{mazur2021target} & 80.64 ($\pm$1.25) &\\
\midrule
\textbf{Multi-Head and Task-Aware Methods} & & \\
\midrule
Synaptic Intelligence$^\dagger$ \cite{zenke2017continual}  & 99.14 ($\pm$0.49) & 94.75 ($\pm$0.49) \\
Deep Generative Replay + distill.$^\dagger$  \cite{shin2017continual} & 99.59 ($\pm$0.40) & 97.51 ($\pm$0.04) \\
Variational Continual Learning$^\dagger$ \cite{nguyen2017variational} & 98.50 ($\pm$1.78) & 96.60 ($\pm$1.34) \\
Elastic Weight Consolidation$^\dagger$ \cite{kirkpatrick2017overcoming}  & 85.48 ($\pm$5.36) & 94.74 ($\pm$0.22) \\
Learning without Forgetting$^\dagger$ \cite{li2017learning} & 99.60 ($\pm$0.13) & 69.84 ($\pm$2.00) \\
Continual Unsupervised Representation Learning \cite{rao2019continual} & 99.10 ($\pm$0.06)& \\
Self-Attention Meta-Learning for CL \cite{sokar2021self} & 97.95 ($\pm$0.07) & \\
Dynamically Expandable Networks \cite{yoon2018lifelong} & 99.26 ($\pm$0.01) & \\
\bottomrule
\end{tabularx}
\caption{Continual learning results with additional task-aware methods in the split MNIST (S-MNIST) and permuted MNIST (P-MNIST) benchmark compared to current CL methods. Results were averaged over ten random seeds with the standard deviation given in parenthesis. Results on algorithms marked with $\dagger$ were taken from \cite{van2018generative}, others from their original work.}
\label{tab:appmnistresults}
\end{table}

\subsection{CIFAR-10 Experiments}
The VAE encoder consisted of five convolutional layers with stride 4 with two experts, each with 8, 16, 32, 64, and 128 units, followed by two dense units with two experts, each with 256 units. The latent variable has 128 dimensions, which we model by two dense layers: one with 128 units for the mean and one with 128 units for the log-variance. The dense layer modeling the mean has three experts, the layer for the log-variance one expert. We assume isotropic Gaussian distributions as priors and posteriors over the latent variable, which allows us to compute the $\DKL$ if closed form. The decoder mirrors the encoder and has two dense layers followed by 5 de-convolutional layers with stride 4  (the last layer has stride 3). All hidden layers use a leaky ReLU activation function \cite{maas2013rectifier}. The VAE output activation function is a sigmoid and we trained it using a binary cross-entropy loss between the normalized pixel values of the original and the reconstructed images. We used no other regularization methods on the VAE. We used 10.000 generated samples after each task.

The classifier architecture is similar to the encoder architecture. We used five convolutional layers, followed by two dense layers. All layers used two experts. The convolutional layers have 8, 16, 32, 64, and 128 units per experts, while the dense layers both have 256 units per layer. We used leaky ReLU as an activation function for the hidden layers and softmax for the output layer. We trained the classifier using a binary cross-entropy loss between the true and the predicted label. We trained with batch size 256 for 1000 epochs using the Adam optimizer with a learning rate of $3*10^{-4}$.

\begin{table}[t!]
\small
\setlength{\tabcolsep}{2pt} 
\centering
\begin{tabularx}{\columnwidth}{*{3}l}
\toprule
\textbf{Baselines} &  \textbf{Split-CIFAR-10} & \textbf{CIFAR-100} \\
\midrule
Conv. Neural Network &  66.62 ($\pm$1.06)  & 19.80 ($\pm$0.19) \\
Offline re-training + task oracle &  80.42 ($\pm$0.95)& 52.30 ($\pm$0.02) \\
\midrule
\textbf{Single-Head and Task-Agnostic Methods} & & \\
\midrule
Hierarchical VCL (ours) & 78.41 ($\pm$1.18) & 33.10 ($\pm$0.62) \\
Hierarchical VCL w/ GR (ours) & 81.00 ($\pm$1.15) & 37.20 ($\pm$0.52) \\
Continual Learning with Dual Regularizations \cite{han2021continual} & 86.72 ($\pm$0.30) & 25.62 ($\pm$0.22) \\
Natural Continual Learning \cite{kao2021natural} & & 38.79 ($\pm$0.24) \\ 
Target Layer Regularization \cite{mazur2021target} & 74.89 ($\pm$0.61 & \\
Memory Aware Synapses \cite{he2022online} & 73.50 ($\pm$1.54)  & \\
\midrule
\textbf{Multi-Head and Task-Aware Methods} & & \\
\midrule
Gradient Episodic Memory \cite{lopez2017gradient} & 79.10 ($\pm$1.60) & 40.60 ($\pm$1.90)  \\
Meta-Consolidation for Continual Learning \cite{kj2020meta} & 82.90 ($\pm$1.20) & 43.50 ($\pm$0.60)\\
Contrastive CL with Feature Propagation \cite{han2021contrastive} & 86.33 ($\pm$1.47) & 65.19 ($\pm$0.65)\\
Hindsight Anchor Learning \cite{chaudhry2021using} & 75.19 ($\pm$2.57) & 47.88 ($\pm$2.76)  \\
Synaptic Intelligence \cite{zenke2017continual} & 63.31 ($\pm$3.79) & 36.33 ($\pm$4.23)\\
Averaged Gradient Episodic Memory \cite{chaudhry2018efficient} & 74.07 ($\pm$0.76) & 46.88 ($\pm$1.81)\\
\bottomrule
\end{tabularx}
\caption{Continual learning results with additional task-aware methods in the split CIFAR-10 and the split CIFAR-100 benchmark compared to current CL methods. Results were averaged over ten random seeds with the standard deviation given in the parenthesis. We report results of other methods as given in their original studies.}
\label{tab:appcifarresults}
\end{table}

\subsection{Reinforcement Learning Experiment Details}
\label{app:crlexperiments}
Each task was trained for one million time steps. We use two layer networks (actor and critics) with 64 units per layer. Each layer has four experts followed by leaky ReLU \cite{maas2013rectifier} activation functions. Each We set each SAC related hyper-parameter as proposed in the original publication \cite{haarnoja2018soft}. We update every 5000 environment steps with a batch size of 256. We use a Prioritized Replay Buffer \cite{schaul2015prioritized} with 1 Million sample capacity.
For UCL \cite{ahn2019uncertainty}, we used the implementation provided by the authors for our experiments and use the hyper-parameters suggested in the publication. Note that the UCL implementation rests on a PPO \cite{Schulman2017} backbone. Our CRL experiments do not use any form of additional replay (except for the replay buffer used by SAC).

\end{appendices}

\end{document}